\documentclass[conference]{IEEEtran}
\IEEEoverridecommandlockouts
\usepackage{cite}
\usepackage{amsmath,amssymb,amsfonts}
\usepackage{algorithmic}
\usepackage{graphicx}
\usepackage{textcomp}
\usepackage{xcolor}
\def\BibTeX{{\rm B\kern-.05em{\sc i\kern-.025em b}\kern-.08em
    T\kern-.1667em\lower.7ex\hbox{E}\kern-.125emX}}
    
\usepackage{algorithm}
\usepackage{subcaption}
\usepackage{multirow}

\begin{document}

\title{Heterogeneous Graph Sparsification for Efficient Representation Learning
\thanks{Chun Jiang Zhu is the corresponding author and was supported by University of North Carolina Greensboro start-up funds. This work was partially supported by NSF grant IIS-1718738, and NIH grants R01DA051922-02, 5R01MH119678-02, and 5K02DA043063-05 to J. Bi.}
}

\author{
    \IEEEauthorblockN{Chandan Chunduru\IEEEauthorrefmark{1}, Chun Jiang Zhu\IEEEauthorrefmark{1}, Blake Gains\IEEEauthorrefmark{2}, Jinbo Bi\IEEEauthorrefmark{2}}
    \IEEEauthorblockA{\IEEEauthorrefmark{1}Department of Computer Science, University of North Carolina Greensboro, Greensboro, NC, USA
    \\\{c\_chunduru, chunjiang.zhu\}@uncg.edu}
    \IEEEauthorblockA{\IEEEauthorrefmark{2}Department of Computer Science and Engineering, University of Connecticut, Storrs, CT, USA
    \\\{blake, jinbo.bi\}@uconn.edu}
}

\IEEEoverridecommandlockouts
\IEEEpubid{\makebox[\columnwidth]{978-1-6654-6819-0/22/\$31.00~\copyright2022 IEEE \hfill} \hspace{\columnsep}\makebox[\columnwidth]{ }}

\maketitle

\IEEEpubidadjcol

\begin{abstract}
Graph sparsification is a powerful tool to approximate an arbitrary graph and has been used in machine learning over homogeneous graphs. In heterogeneous graphs such as knowledge graphs, however, sparsification has not been systematically exploited to improve efficiency of learning tasks. In this work, we initiate the study on heterogeneous graph sparsification and develop sampling-based algorithms for constructing sparsifiers that are provably sparse and preserve important information in the original graphs. We have performed extensive experiments to confirm that the proposed method can improve time and space complexities of representation learning while achieving comparable, or even better performance in subsequent graph learning tasks based on the learned embedding.
\end{abstract}


\section{Introduction}
Heterogeneous graphs (which subsume knowledge graphs) can model complex relationships among multiple types of real-world entities and have been widely used in studying biological and biomedical data. In a heterogeneous graph (\emph{e.g.}, \cite{fan2019metapath,zhang2019heterogeneous,nicholson2020constructing}), there can be multiple types of nodes (\emph{e.g.}, some for compounds and others for targets, gene or diseases) and edges (\emph{e.g.}, a compound treats a disease, a compound resembles another compound, or a gene interacts with another gene), making discovery of deep insights (\emph{e.g.}, perform drug repurposing or identify drug-target interactions) become possible. Graph representation learning aims to construct low-dimensional embeddings (vectors) for all nodes in a graph, and has been a standard approach to further knowledge discovery and development on graphs. For instance, both predicting biomedical network links based on their structural and feature information and clustering for drug-target interaction and drug discovery are based on high-quality heterogeneous network representation. Recently, various heterogeneous graph representation learning methods have been proposed, \emph{e.g.}, random-walk based methods Metapath2Vec \cite{dong2017metapath2vec} and HIN2Vec \cite{fu2017hin2vec}, message-passing based graph neural networks HAN \cite{wang2019heterogeneous} and HGT \cite{hu2020heterogeneous}, and relation learning based methods TransE \cite{bordes2013translating}, ComplEx \cite{trouillon2016complex}, and RotatE \cite{sun2019rotate}. For a systematic review on heterogeneous graph embedding, readers are referred to a good survey \cite{yang2020heterogeneous}.

However, modern heterogeneous graphs can be of very large size in terms of the number of nodes or the number of edges. For instance, the benchmarking knowledge graph Freebase has over $12$ millions nodes across $8$ node types and about $63$ millions edges of $36$ edge types \cite{yang2020heterogeneous}. Processing such large-scale heterogeneous graphs imposes high computational and storage demands. We identify two directions to meet the requirements: one approach is to develop efficient heterogeneous embedding algorithms. But this direction may not offset the effects of ever growing big heterogeneous graphs. The other is to sparsify/compress big heterogeneous graphs while approximating some important network property and not significantly impacting the quality of graph embedding. In this work, we focus on the latter topic of sparsifying heterogeneous graphs.

Graph sparsification deals with approximating an arbitrary graph by a sparse graph that usually has a linear number of edges in the number of vertices of the original graph while preserving some important property.
Several notions of graph sparsification have been proposed, \emph{e.g.}, graph spanners \cite{PS89} that approximate the shortest path distance between every pair of vertices, cut sparsifiers \cite{BK96} that approximately preserve all graph cut values, and spectral sparsifiers \cite{ST11} that approximate the graph spectrum/eigenvalues. Graph sparsification has been used to improve the computational cost of learning over graphs, \emph{e.g.}, graph semi-supervised learning and spectral clustering \cite{SWT16,ZLB21a,ZLB21b}. Although heterogeneous graphs can have much larger size since heterogeneity allows more nodes or edges to be included in one network, however, there has been no systematic study on sparsification techniques for heterogeneous graphs yet.

In this paper, we for the first time study the problem of heterogeneous graph sparsification and its algorithmic applications in graph representation learning. We propose an efficient algorithm for constructing provably smaller sparsifiers for heterogeneous graphs, which, when feeding into a heterogeneous graph embedding algorithm, can save significant computational and storage cost. However, interestingly, the quality of the learned representations using sparsifiers is comparable or even \emph{better} than those based on the original graphs. We have provided theoretical analysis of the saving in time and space complexities of heterogeneous graph representation learning. Extensive experiments have also been performed to demonstrate that the developed method can improve the efficiency of heterogeneous graphs while not significantly affecting the accuracy of downstream machine learning tasks based on the learned embedding.

{\noindent \bf Related Work.}
Sparsification over homogeneous graphs has been extensively studied, \emph{e.g.}, \cite{PS89,BK96,ST11} and used in machine learning \cite{SWT16,CKL+18,ZSL+19a}. In heterogeneous graphs, however, how to sparsify various types of vertices and edges while preserving graph properties has received limited studies \cite{jiang2021pre,yang2020heterogeneous}. Jiang \emph{et al.} \cite{jiang2021pre} developed self-supervised pre-training methods for heterogeneous graphs with relation-based sparsification for efficient pre-training. But their focus is on designing both the node- and schema-level pre-training tasks to preserve semantic and structural properties, instead of a comprehensive study on the sparsification method.
To the best of our knowledge, this work is the first systematical study on heterogeneous graph sparsification and its use in representation learning.



\section{Heterogeneous Graph Sparsifiers}

In this section, we develop a sampling-based sparsification technique for heterogeneous graphs, analyze its computational and space complexities, and then discuss other potential methods.

{\noindent \bf Notations and Definitions.}
A heterogeneous graph can be defined as $G(V,E,\phi,\pi,X,R)$ where $V$ is the vertex set, $E$ is the edge set, $\phi$ specifies the type for each vertex, $\pi$ assigns the type of each edge, and $X$ and $R$ include the node and edge features respectively. $\phi$, $\pi$, $X$, and $R$ can be omitted from the presentation if they are clear from the context. Let $E_{Out}(u)$ be the set of $u$'s out-edges and $E_{Out}^t(u)$ be the set of $u$'s out-edges of type $t$. Similarly, let $E_{In}(u)$ be the set of $u$'s in-edges and $E_{In}^t(u)$ be the set of $u$'s in-edges of type $t$. Let $|G(V,E)|=|E|$.

{\noindent \bf Motivation.}
Graph sparsification in general/homogeneous graphs has received considerable attention and many sparsification algorithms have been developed. However, they can be separated into two types, \emph{importance} based sampling methods and \emph{heuristic} based sampling methods. In the former, graph edges are sampled according to some important measure. For example, spectral sparsifiers and cut sparsifiers can be obtained by non-uniformly sampling each edge based on its effective resistance and edge connectivities \cite{SS11,FHH+11}, respectively. This type of approaches is often sound and has nice theoretical guarantees. However, such a design for heterogeneous graphs can be highly challenging considering the complexity in defining an importance measure across different types of edges and computing that measure in large-scale heterogeneous graphs. 

In the latter, sampling is performed on all edges independently using only heuristics instead of complex importance measures. For example, $k$-neighbors sparsifiers are constructed by randomly sampling at most $k$ edges for each vertex where $k>0$ is a tuneable parameter. Though simple, heuristic based sparsifiers have performed practically good, \emph{e.g.}, comparable to spectral sparsifiers in machine learning tasks over graphs \cite{SWT16}. In this paper, we will work towards sparsifying heterogeneous graphs using simple yet effective heuristics.

{\noindent \bf Details.}
As a warm-up, we consider the simple method of sampling a fixed number of edges and including in the sparsifier. Since node degrees can be quite different for different nodes, it is possible that a node of low degree has none of its edges sampled and become \emph{isolated} in the sparsifier. Unfortunately, some heterogeneous embedding algorithms do not allow isolated nodes (through reporting errors such as \cite{shi2018aspem,wang2019heterogeneous}) and thus this type of sparsifiers is not robust to embedding algorithms. 
Furthermore, the number of edges of different types for a node can be very different because of their semantic difference. Consider an extreme case: a vertex $u$ with $10K$ edges of type $1$ and only one edge of type $2$. The information carried by the edge of type $2$ is supposed to be greater than that in an individual edge of type $1$. When all edges are considered together, the single edge of type $2$ can be easily sampled away, resulting in information loss.

We propose to perform random sampling within each type of edges and then aggregate all sampled edges as the sparsfiier. In this way, the isolation problem is avoided and the distribution of edge types cannot affect the quality of the constructed sparsifier.
Specifically, we sparsify the neighborhood of each vertex according to ascending order of vertex degree. Since heterogeneous graphs are often directed, we sparsify out-edges and in-edges of a vertex separately. Consider out-edges $E_{Out}(u)$ of vertex $u$. For each edge type $t$ in $E_{Out}(u)$, if the number of edges of type $t$ is at most $k$, we include all of them to the sparsifier $H$. Here $k$ is an input parameter to control the size of the output sparsifier. Otherwise, we first calculate the number of edges in $E^t_{Out}(u)$ that are already included in the current $H$. Note that we process vertices one by one and the processing of some vertex before the current vertex may already add a few edges in $E_{Out}(u)$ to $H$. Let $x$ be the number of such edges, $x=|E_{Out}^t(u)\cap H|$. Then we sample $k-x$ edges from $E_{Out}^t(u)-H$ uniformly at random and include the sampled edges in $H$. Since in heterogeneous graphs there can be many edges without edge weight, we keep the weight of a sampled edge unchanged, if any. The processing for in-edges is symmetric. The algorithms are summarized in Algorithms \ref{alg:graph} and \ref{alg:node1}.

\begin{algorithm}[t]
\small
\caption{Sparsify-Graph}
\renewcommand{\algorithmicrequire}{\textbf{Input:}}
\renewcommand{\algorithmicensure}{\textbf{Output:}}
\begin{algorithmic}[1]
\REQUIRE A graph $G$ and a parameter $k$
\ENSURE A sparsified graph
\STATE $H=\emptyset$;
\STATE Sort vertices in ascending order of degree;
\FOR {each node $u$ in the sorted order}
    \STATE Sparsify-Node-Out($u$, $G$, $H$, $k$);
    \STATE Sparsify-Node-In($u$, $G$, $H$, $k$);
\ENDFOR
\RETURN $(V,H)$;
\end{algorithmic}
\label{alg:graph}
\end{algorithm}

\begin{algorithm}[t]
\small
\caption{Sparsify-Node-Out(-In)}
\renewcommand{\algorithmicrequire}{\textbf{Input:}}
\renewcommand{\algorithmicensure}{\textbf{Output:}}
\begin{algorithmic}[1]
\REQUIRE A node $u$ in a graph $G$, an edge set $H$, and a parameter $k$
\ENSURE Updated $H$
\FOR {each edge type $t$}
    \IF {$|E_{Out}^t(u)|<k$ \COMMENT{respectively, $|E_{In}^t(u)|<k$}}
        \STATE Include $E_{Out}^t(u)$ into $H$;
    \ELSE
        \STATE Compute $x=|E_{Out}^t(u)\cap H|$; \COMMENT{$x=|E_{In}^t(u)\cap H|$;}
        \STATE Sample $k-x$ edges from $E_{Out}^t(u)-H$ uniformly at random; let the sampled edges be $S$; \{Sample $k-x$ edges from $E_{In}^t(u)-H$;\} 
        \STATE Include $S$ into $H$;
    \ENDIF
\ENDFOR
\RETURN $H$;
\end{algorithmic}
\label{alg:node1}
\end{algorithm}

The computational and space complexities of Algorithm \ref{alg:graph} are $O(ktn)$ and the size (number of edges) of the resulting sparsifier is also $O(ktn)$, where $t$ is the total number of edge types. Let the computational complexity of a machine learning task in a heterogeneous graph $G$ be $T'(G)$. It is generally believed to be a function of the number of edges $m=|E|$, \emph{i.e.}, $T'(G)=T(m)$. Although there are few studies on the time complexity of learning in heterogeneous graphs, in homogeneous graphs, the time complexities of graph-based learning, \emph{e.g.}, supervised learning such as Laplacian smoothing, semi-supervised learning, and spectral clustering, have shown to be linear in the number of edges \cite{CKL+18,ZSL+19a}. We believe that similar relations appear in heterogeneous graphs. Based on this assumption, our approach can reduce the computational cost from $T(m)$ to $O(ktn)+T(ktn)$. The improvement is significant because $m\gg n$ usually holds for heterogeneous graphs, and $k$ and $t$ are only a small value. This analysis is applicable to the space complexity as well. We will confirm the improvement through empirical studies in Section \ref{sec:experiment}.

{\noindent \bf Discussion.}
We develop another method that also does not suffer from the isolation problem. The principle is to first randomly select, for every edge type $t$ of vertex $u$, one edge from $E^t_{Out}(u)$ (and $E^t_{In}(u)$, respectively) to include in the sparsifier. Let the number of edges included be $y$. We then randomly keep $k-y$ edges out of all $u$'s edges, regardless of edge types. This method guarantees that each edge type $t$ of $u$'s edges has at least one sampled edges and thus does not lose important information. This algorithm constructs sparsifiers of size $O(\max(k,t)\cdot n)$, which is smaller than Alg. \ref{alg:graph} for the same value of $k$. Consider sparsifiers of similar size produced by both methods, their empirical performance in subsequent machine learning tasks are comparable. As will be shown in Section \ref{sec:experiment}, sparsifiers produced by Alg. \ref{alg:graph} generally performs better in the link prediction task while sparsifiers output by this method works better in node classification, although the margins are relatively small.


\section{Experiments}
\label{sec:experiment}

In this section, we perform extensive experiments to evaluate the proposed algorithms for sparsifying heterogeneous graphs and performance of the constructed sparsifiers in representation learning.
The algorithms were implemented in Python, and all experiments were performed using 3 cores in a compute node, which is equipped with 24 cores, 2.50 GHz Intel processors, and 256GB RAM.

{\noindent \bf Experimental Setting.}
We use three different real-world datasets provided by the benchmark of \cite{yang2020heterogeneous} and their statistics are summarized in Table \ref{tab:datasets}. \emph{PubMed} is a network of genes, diseases, chemicals, and species from NIH PubMed project (not the traditional publication network). \emph{Yelp} is a review network of businesses, users, locations, and reviews. \emph{Freebase} is a knowledge base of books, films, music, sports, people and so on. Since this is the first work on heterogeneous graph sparsification for representation learning, we also implement the algorithm in the discussion for comparison, which is denoted as \emph{All-Types} (\emph{ALT}). 

The tested graphs are first processed by a sparsification algorithm to get sparsified graphs, which are then fed into a representation learning algorithm to get node embedding. To show the robustness of the proposed sparsification, five diverse algorithms including random-walk based \emph{HIN2Vec} \cite{fu2017hin2vec} and \emph{AspEm} \cite{shi2018aspem}, message-passing based \emph{HAN} \cite{wang2019heterogeneous} and \emph{HGT} \cite{hu2020heterogeneous}, and relation learning based \emph{ComplEx} \cite{trouillon2016complex}, are employed to generate the embedding. Finally, \emph{link prediction} and \emph{node classification} tasks are performed based on the learned embedding following experimental setup in the benchmark \cite{yang2020heterogeneous}. In link prediction, $20\%$ links are held out and the remaining $80\%$ links are used as the training set. A linear Support Vector Machine (LinearSVC) \cite{fan2008liblinear} is trained on the training links. For node classification, we train a LinearSVC based on the learned embedding on $80\%$ of the labeled nodes and then predict on the remaining $20\%$. 

The efficiency for graph learning tasks are measured by the \emph{training time} in seconds and \emph{memory consumption} in term of the maximal residual size during model training. The effectiveness of link prediction is measured by the \emph{AUC} (area under the ROC curve) and \emph{MRR} (mean reciprocal rank). The effectiveness of node classification is measured by \emph{macro-F1} (across all labels) and \emph{micro-F1} (across all nodes). These measures are averaged over \emph{five} runs and reported together with their standard deviation.

\begin{table}[t]
\center
\small
\begin{tabular}{|c|c|c|c|c|c|}
\hline
Dataset & $n$ & $m$ & $m/n$ & {\footnotesize Node Types} & $t$ \\
\hline
\emph{PubMed} & $63,109$ & $236,458$ & $3.7$ & $4$ & $10$ \\
\emph{Freebase} & $12,164,758$ & $62,058,902$ & $5.1$ & $8$ & $36$ \\
\emph{Yelp} & $82,465$ & $16,274,179$ & $197$ & $4$ & $4$ \\
\hline
\end{tabular}
\caption{Statistics of the Tested Real-world Networks}
\label{tab:datasets}
\end{table}

\begin{table}[t]
\center
\small
\begin{tabular}{|c|c|c|c|c|}
\hline
$k$ & \emph{PubMed} & \emph{Freebase} & $k$ & \emph{Yelp} \\
\hline
$1$ & $0.45$ & $0.39$ & $20$ & $0.15$ \\ \hline 
$2$ & $0.59$ & $0.55$ & $50$ & $0.28$ \\ \hline
$3$ & $0.67$ & $0.65$ & $100$ & $0.42$ \\ \hline
$5$ & $0.75$ & $0.73$ & $200$ & $0.59$ \\ \hline
$10$ & $0.86$ & $0.81$ & $500$ & $0.8$ \\ \hline
\end{tabular}
\caption{The Sparsification Ratio ($|H|/|G|)$}
\label{tab:sparsified}
\vspace{-0.2in}
\end{table}

\begin{figure*}[t]
     \center
     \begin{subfigure}[t]{0.32\linewidth}
        \centering\includegraphics[width=\linewidth]{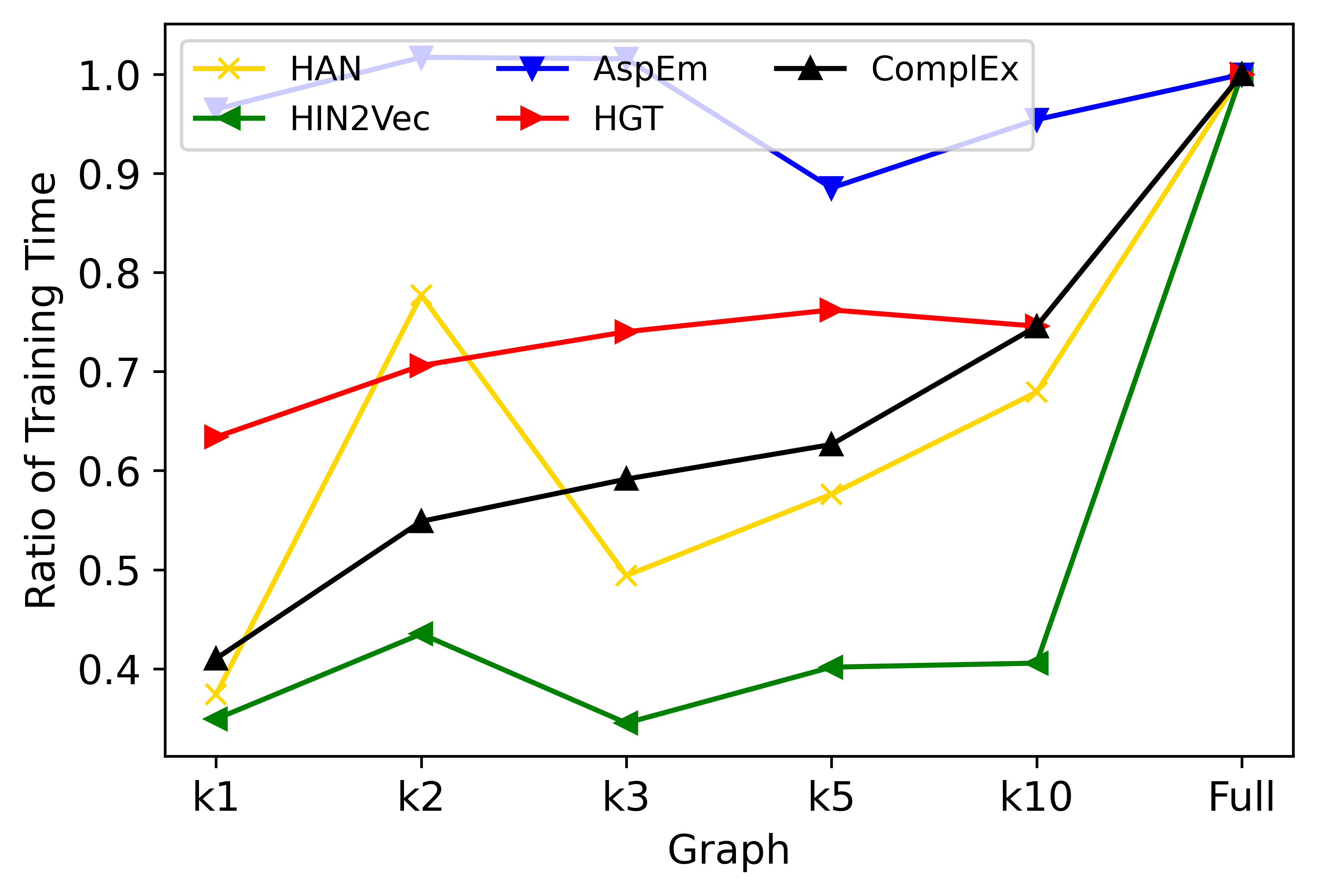}
        \vspace{-0.2in}
        \caption{\small Training Time in \emph{PubMed}}
        \label{fig:time-pubmed}
     \end{subfigure}
     \hspace{\fill}
     \begin{subfigure}[t]{0.32\linewidth}
        \centering\includegraphics[width=\linewidth]{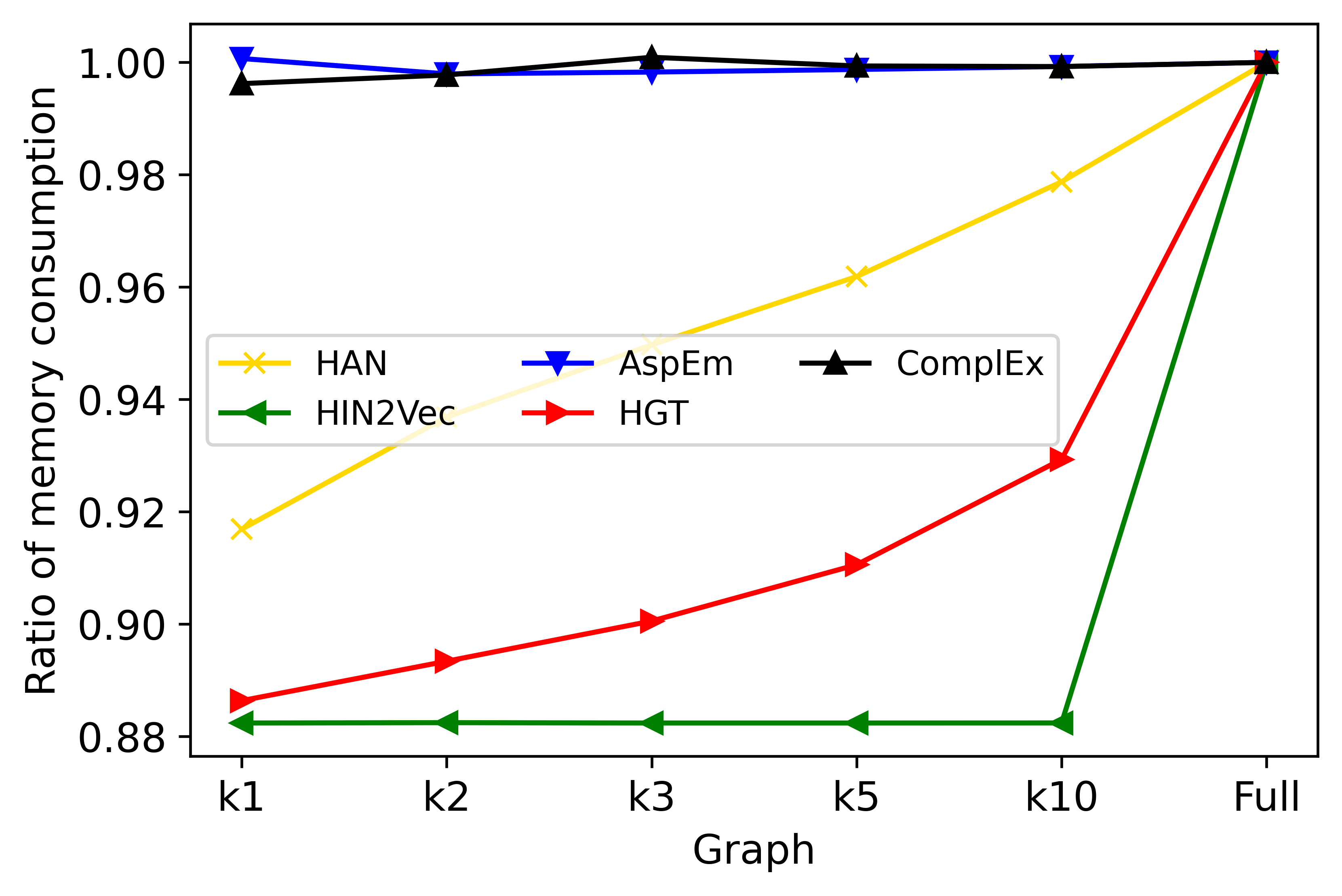}
        \vspace{-0.2in}
        \caption{\small Memory Consumption in \emph{PubMed}}
        \label{fig:ram-pubmed}
     \end{subfigure}
     \hspace{\fill}
     \begin{subfigure}[t]{0.32\linewidth}
        \centering\includegraphics[width=\linewidth]{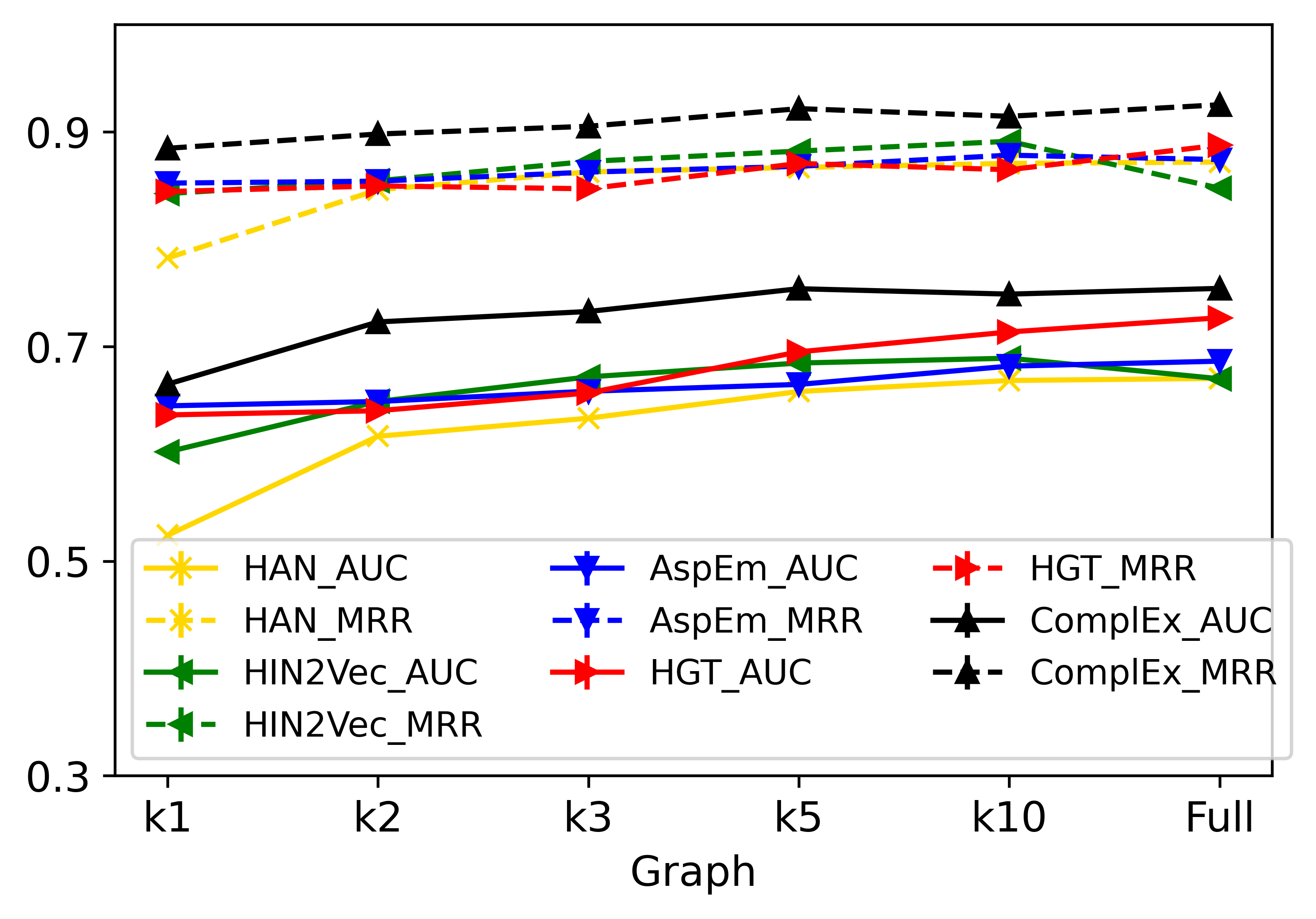}
        \vspace{-0.2in}
        \caption{\small \emph{AUC} \& \emph{MRR} in \emph{PubMed}}
        \label{fig:auc-pubmed}
     \end{subfigure}
     \hspace{\fill}
     \begin{subfigure}[t]{0.32\linewidth}
        \centering\includegraphics[width=\linewidth]{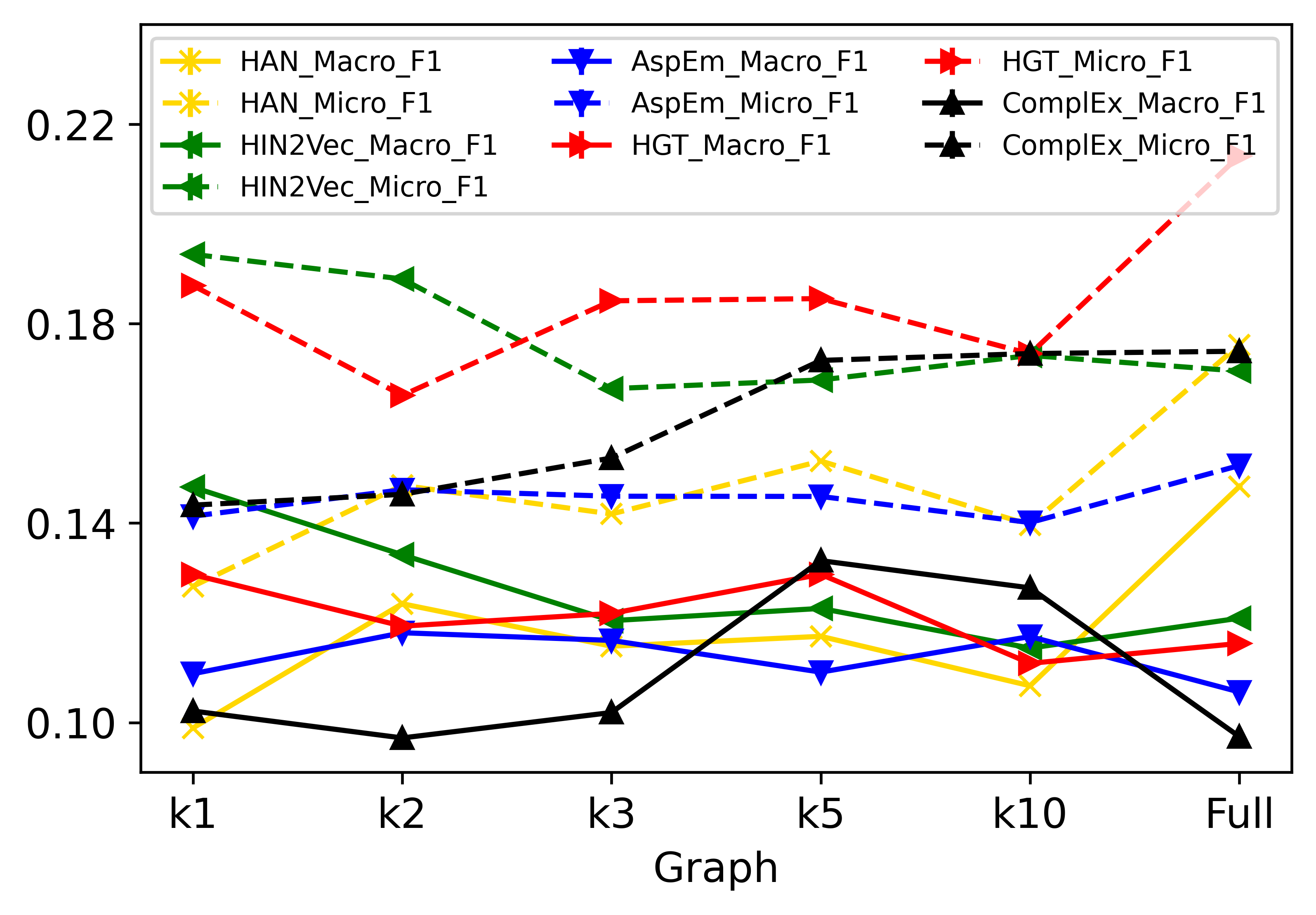}
        \vspace{-0.2in}
        \caption{\small Macro-F1 \& Micro-F1 in \emph{PubMed}}
        \label{fig:f1-pubmed}
     \end{subfigure}
     \hspace{\fill}
     \begin{subfigure}[t]{0.32\linewidth}
        \centering\includegraphics[width=\linewidth]{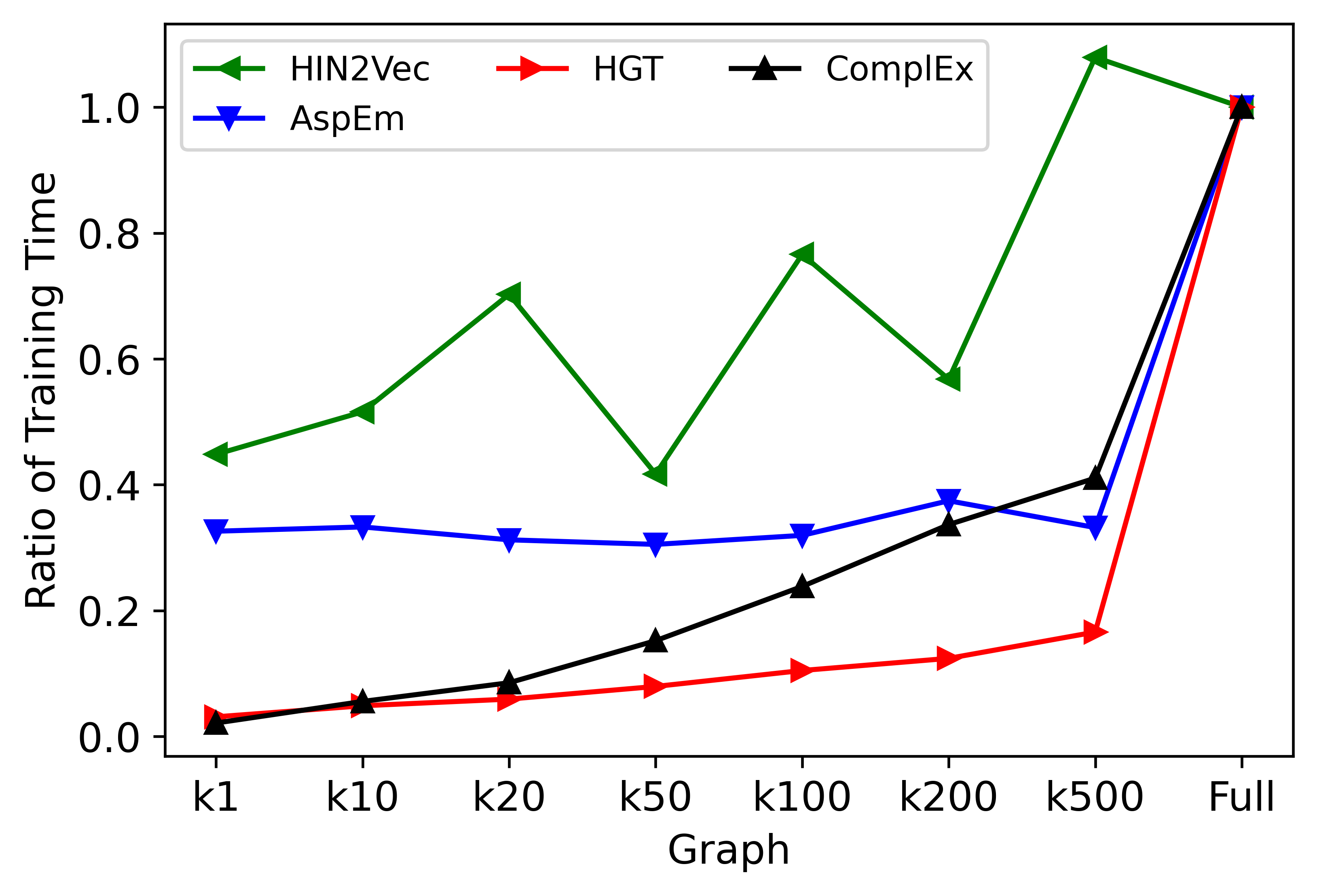}
        \vspace{-0.2in}
        \caption{\small Training Time in \emph{Yelp}}
        \label{fig:time-yelp}
     \end{subfigure}
     \hspace{\fill}
     \begin{subfigure}[t]{0.32\linewidth}
        \centering\includegraphics[width=\linewidth]{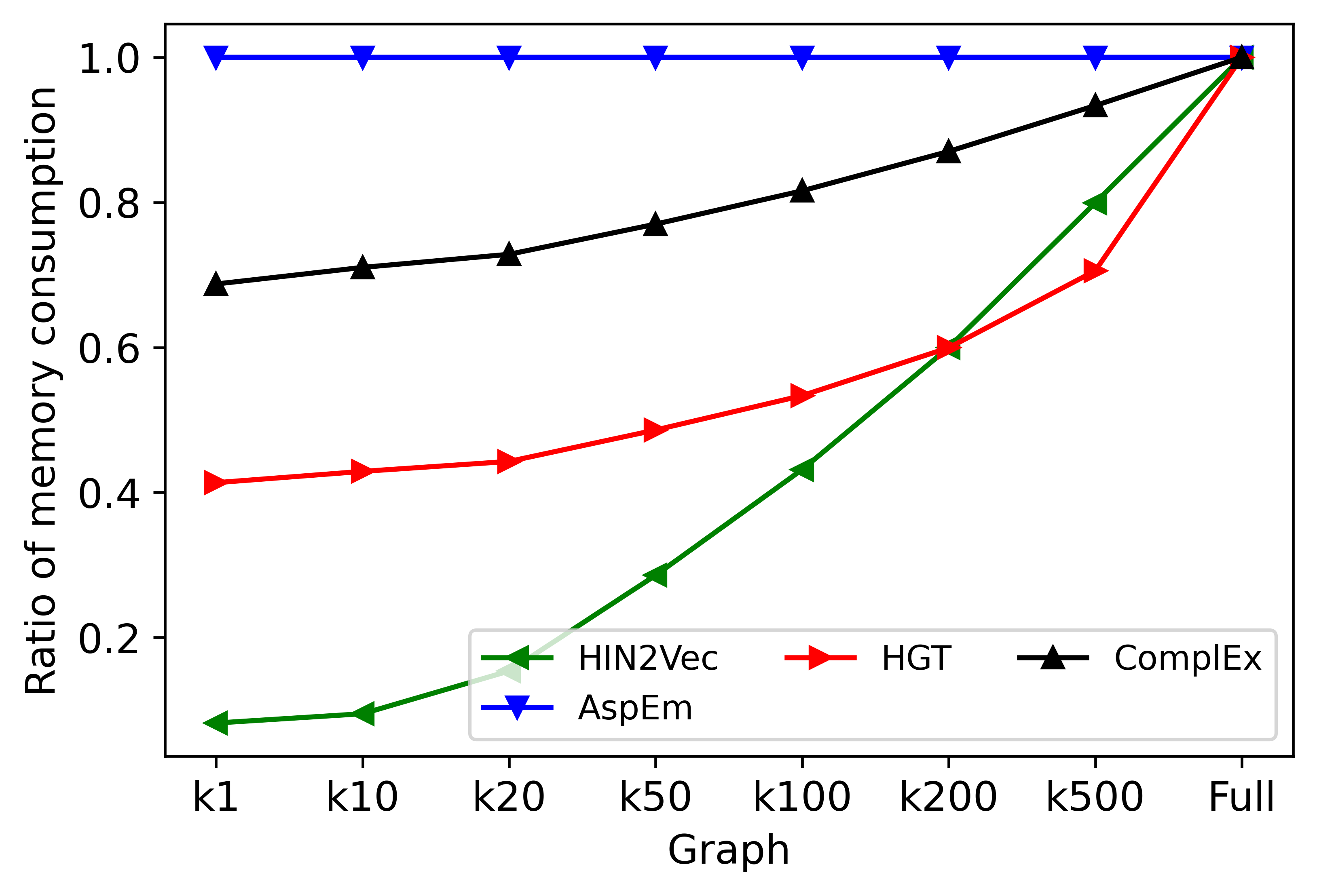}
        \vspace{-0.2in}
        \caption{\small Memory Consumption in \emph{Yelp}}
        \label{fig:ram-yelp}
     \end{subfigure}
     \hspace{\fill}
     \begin{subfigure}[t]{0.32\linewidth}
        \centering\includegraphics[width=\linewidth]{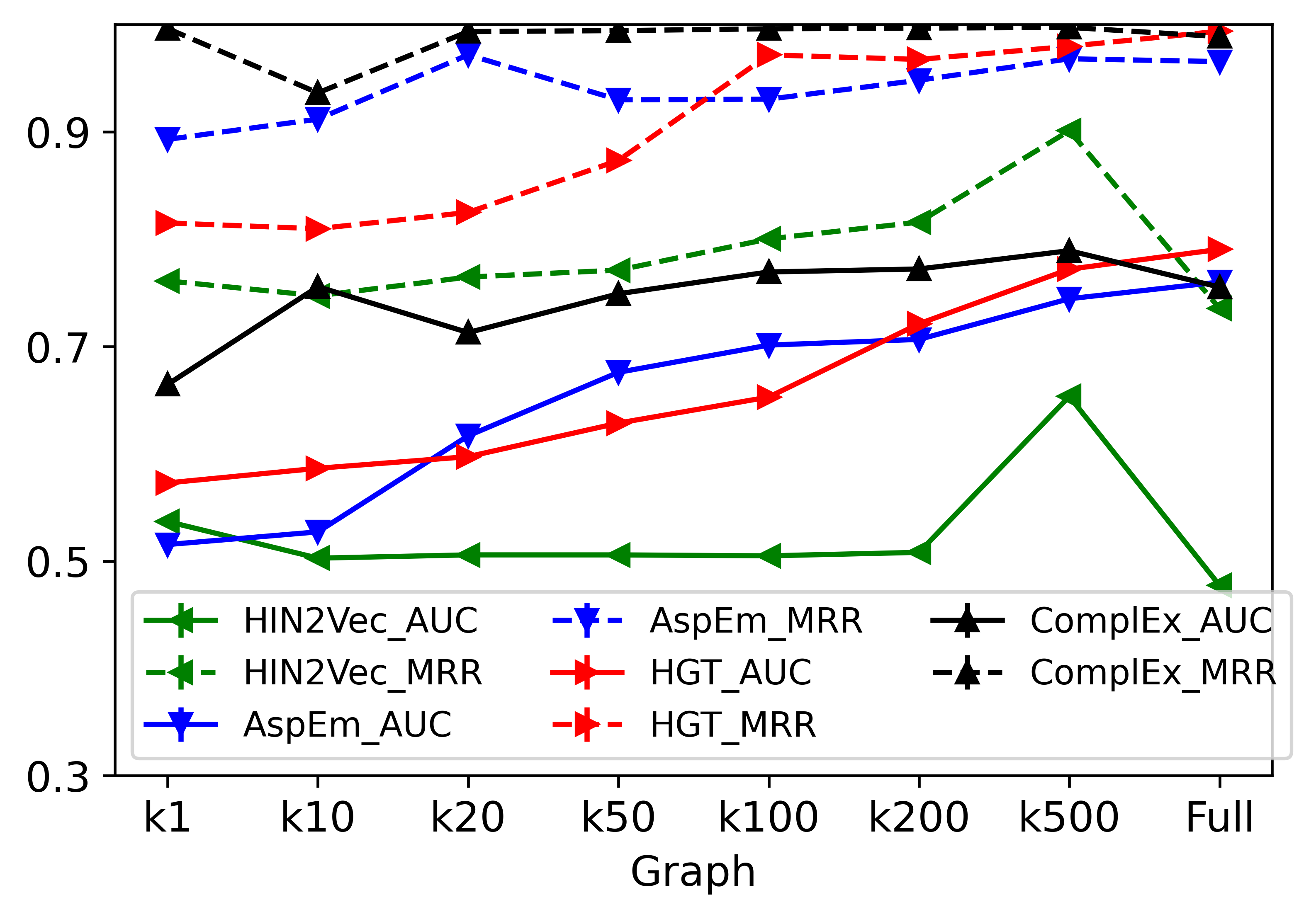}
        \vspace{-0.2in}
        \caption{\small \emph{AUC} \& \emph{MRR} in \emph{Yelp}}
        \label{fig:auc-yelp}
     \end{subfigure}
     \hspace{\fill}
     \begin{subfigure}[t]{0.32\linewidth}
        \centering\includegraphics[width=\linewidth]{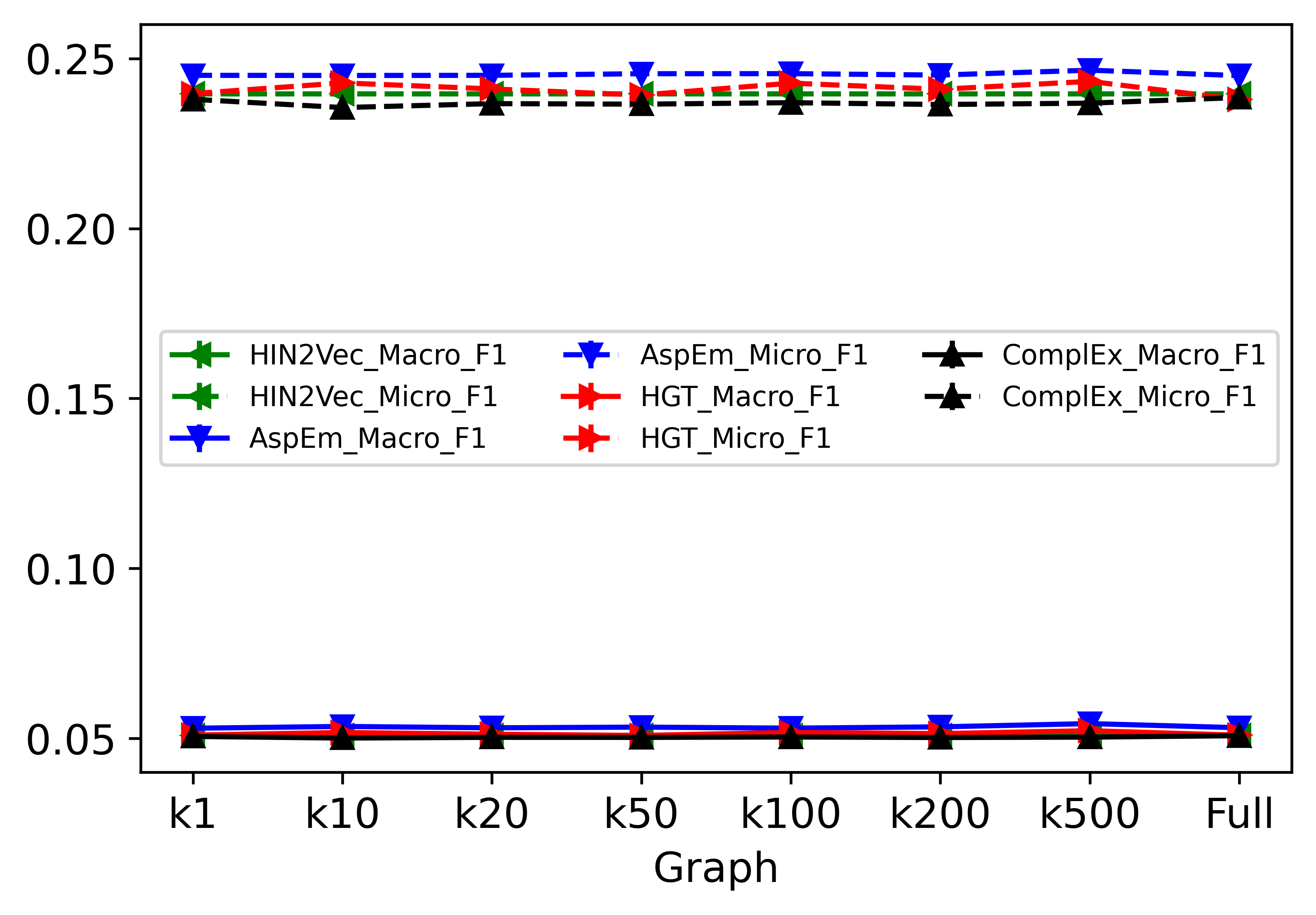}
        \vspace{-0.2in}
        \caption{\small Macro-F1 \& Micro-F1 in \emph{Yelp}}
        \label{fig:f1-yelp}
     \end{subfigure}
     \hspace{\fill}
     \begin{subfigure}[t]{0.32\linewidth}
        \centering\includegraphics[width=\linewidth]{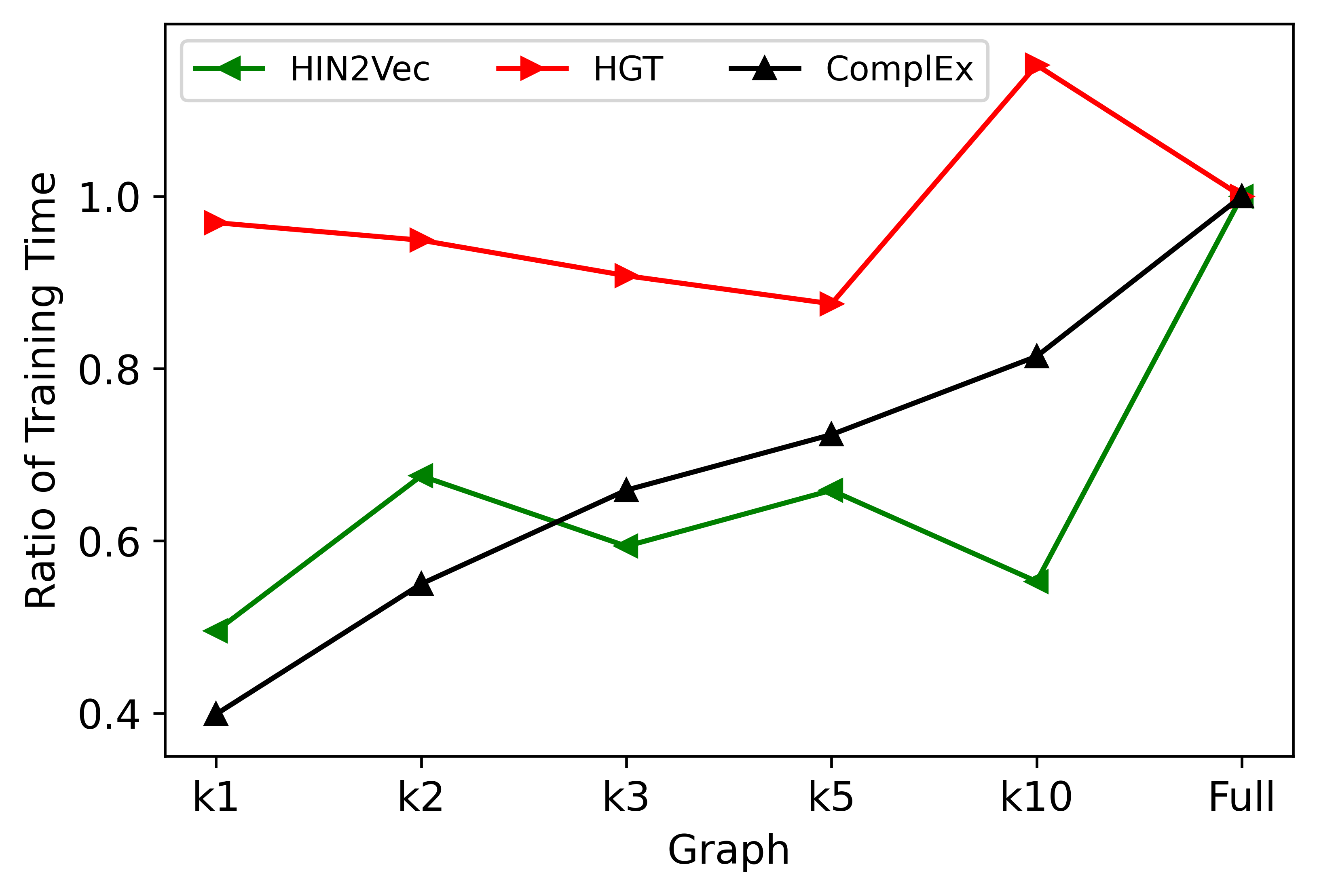}
        \vspace{-0.2in}
        \caption{\small Training Time in \emph{Freebase}}
        \label{fig:time-freebase}
     \end{subfigure}
     \hspace{\fill}
     \begin{subfigure}[t]{0.32\linewidth}
        \centering\includegraphics[width=\linewidth]{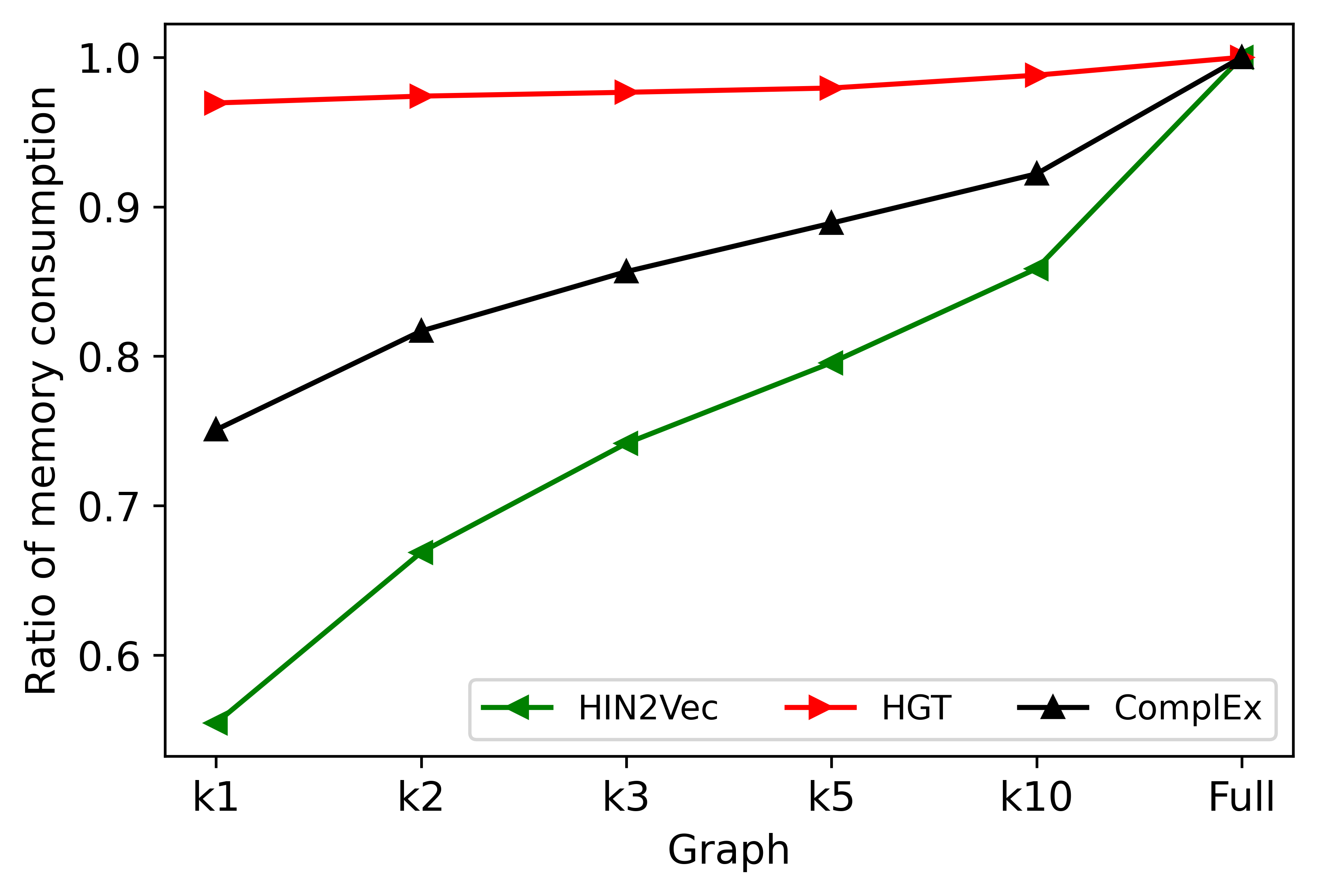}
        \vspace{-0.2in}
        \caption{\small Memory Consumption in \emph{Freebase}}
        \label{fig:ram-freebase}
     \end{subfigure}
     \hspace{\fill}
     \begin{subfigure}[t]{0.32\linewidth}
        \centering\includegraphics[width=\linewidth]{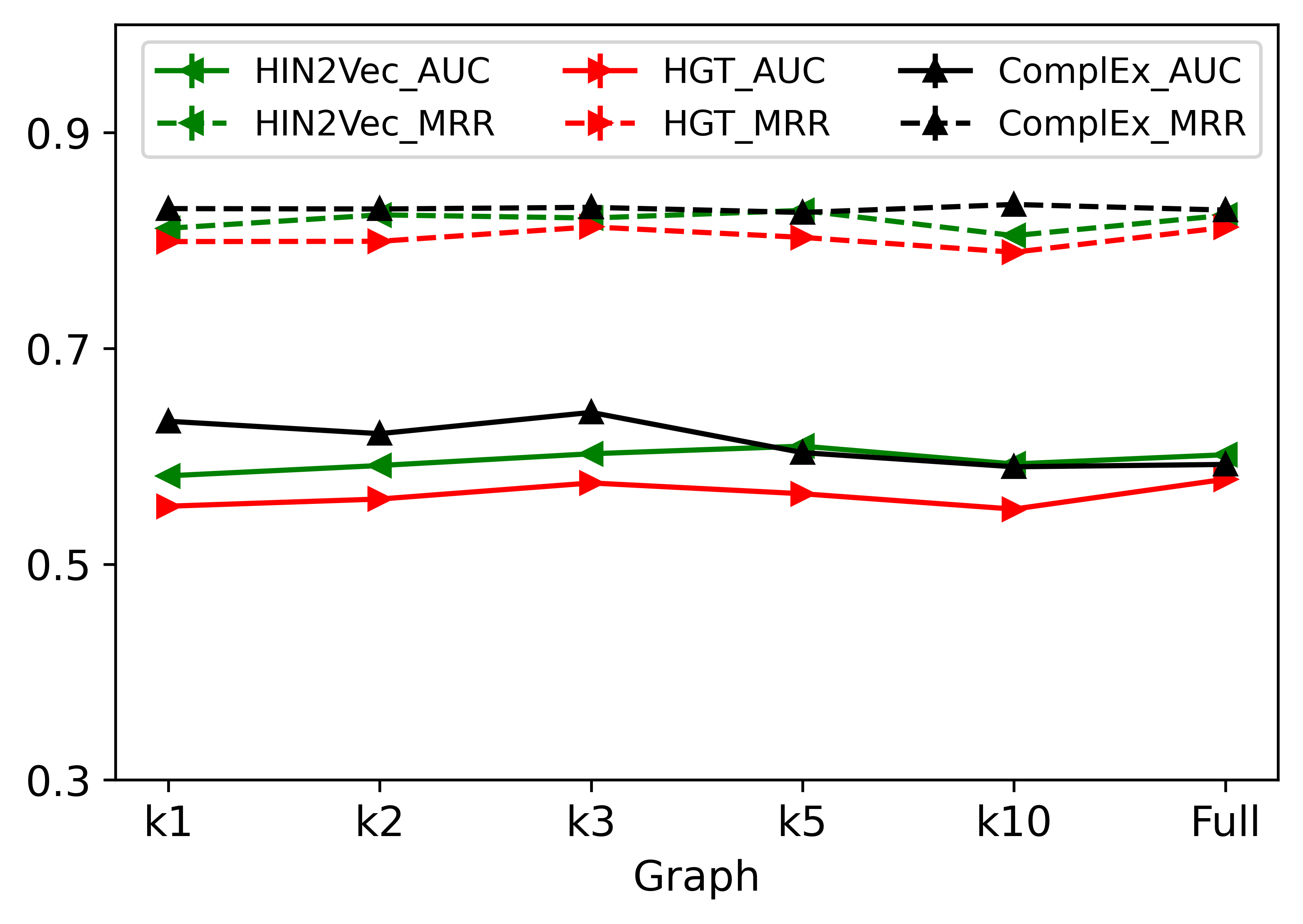}
        \vspace{-0.2in}
        \caption{\small \emph{AUC} \& \emph{MRR} in \emph{Freebase}}
        \label{fig:auc-freebase}
     \end{subfigure}
     \hspace{\fill}
     \begin{subfigure}[t]{0.32\linewidth}
        \centering\includegraphics[width=\linewidth]{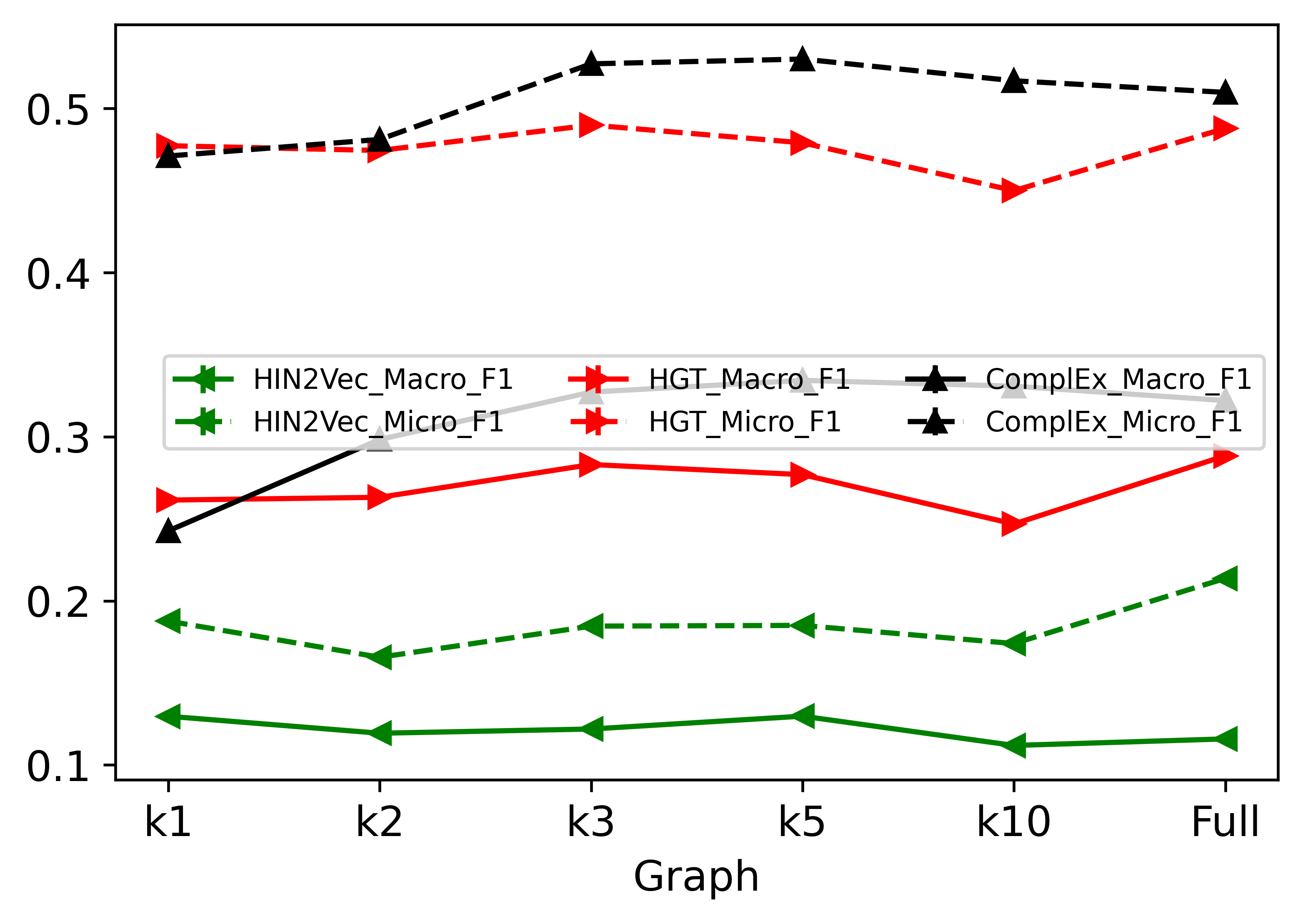}
        \vspace{-0.2in}
        \caption{\small Macro-F1 \& Micro-F1 in \emph{Freebase}}
        \label{fig:f1-freebase}
     \end{subfigure}
     \caption{\small Performance in Different Datasets Under the Baseline Setting}
     \label{fig:results}
     \vskip -0.1in
\end{figure*}

\begin{figure*}[t]
     \center
     \begin{subfigure}[t]{0.32\linewidth}
        \centering\includegraphics[width=\linewidth]{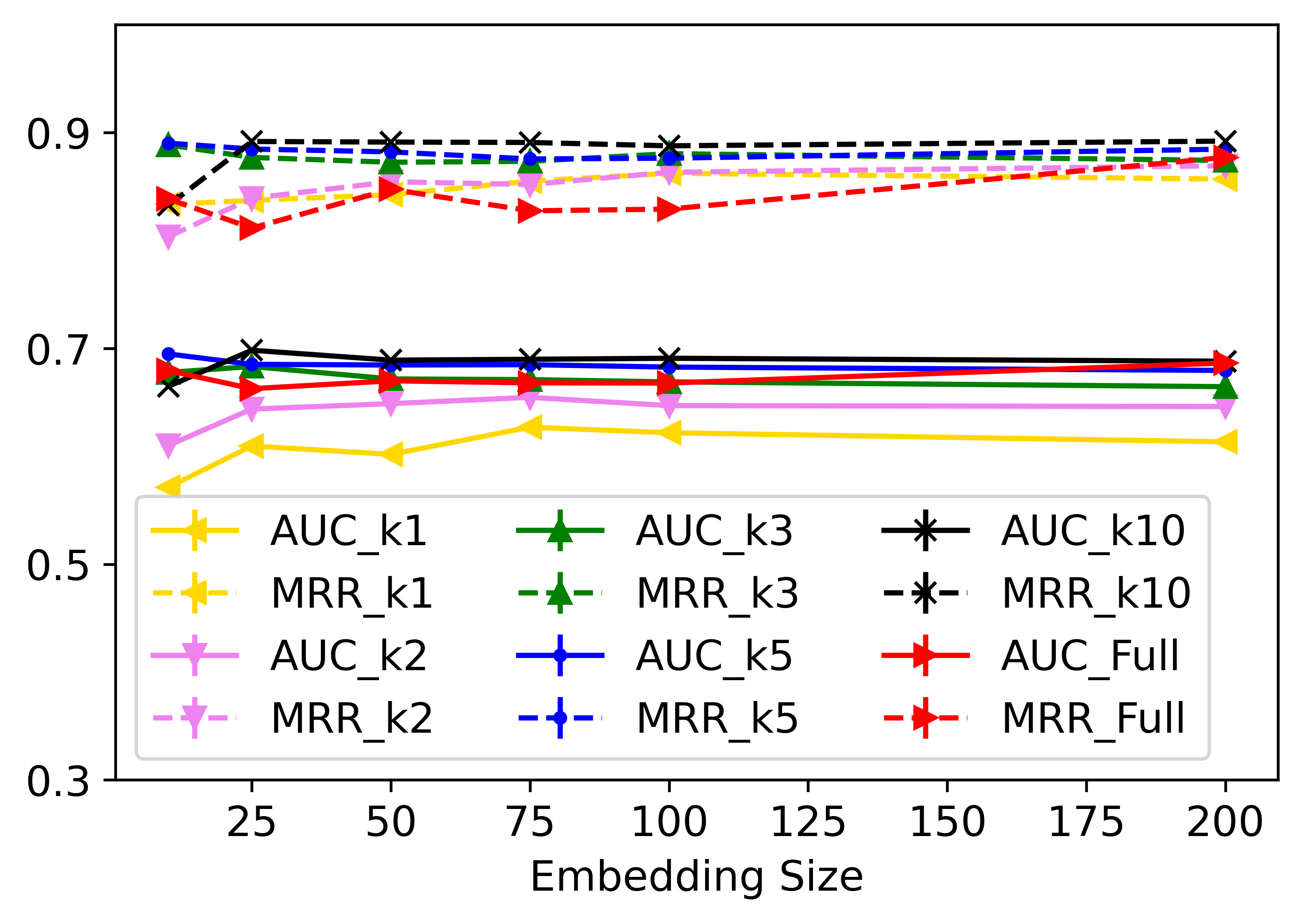}
        \vspace{-0.2in}
        \caption{\small \emph{HIN2Vec}}
        \label{fig:HIN2Vec}
     \end{subfigure}
     \hspace{\fill}
     \begin{subfigure}[t]{0.32\linewidth}
        \centering\includegraphics[width=\linewidth]{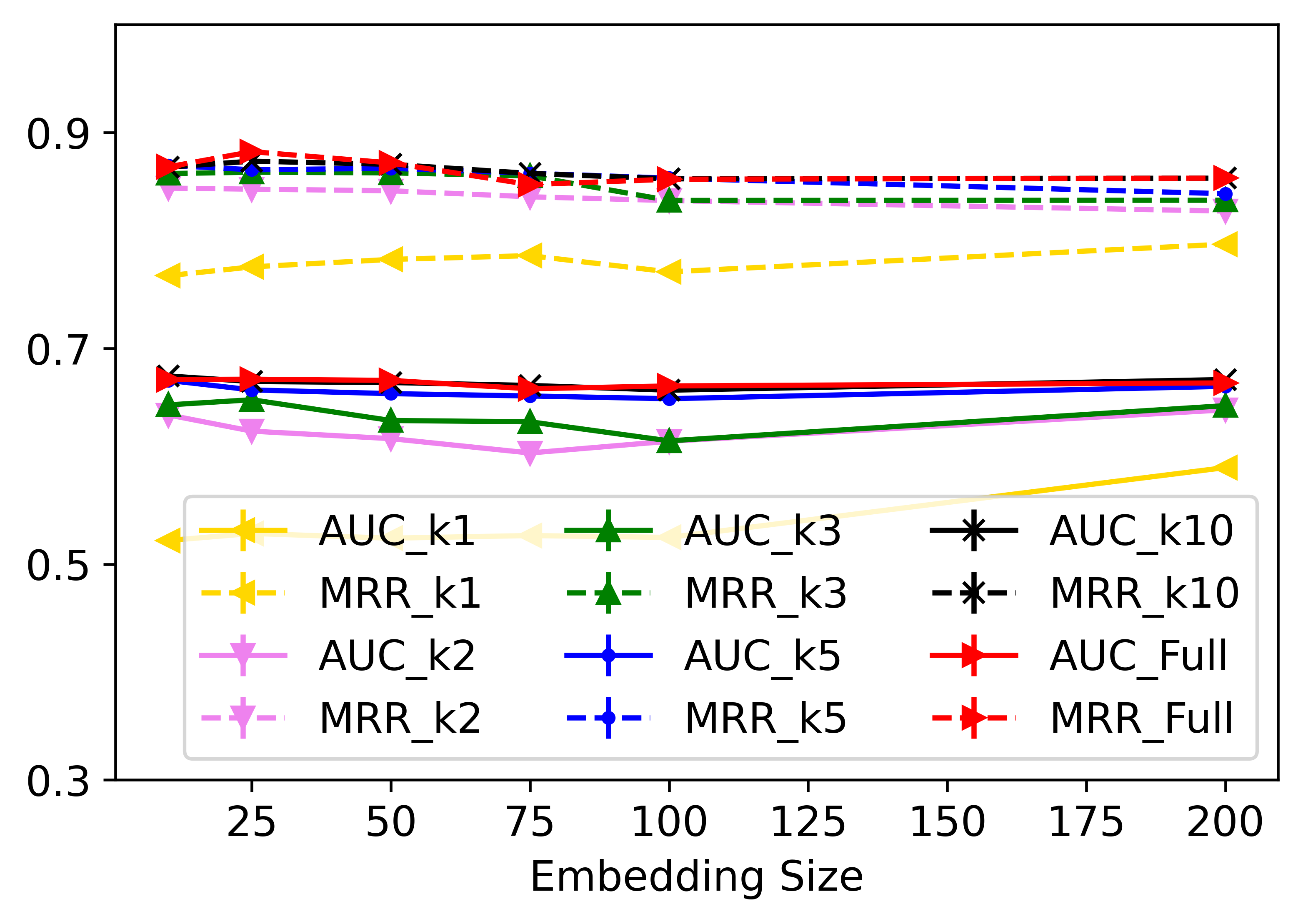}
        \vspace{-0.2in}
        \caption{\small \emph{HAN}}
        \label{fig:HAN}
     \end{subfigure}
     \hspace{\fill}
     \begin{subfigure}[t]{0.32\linewidth}
        \centering\includegraphics[width=\linewidth]{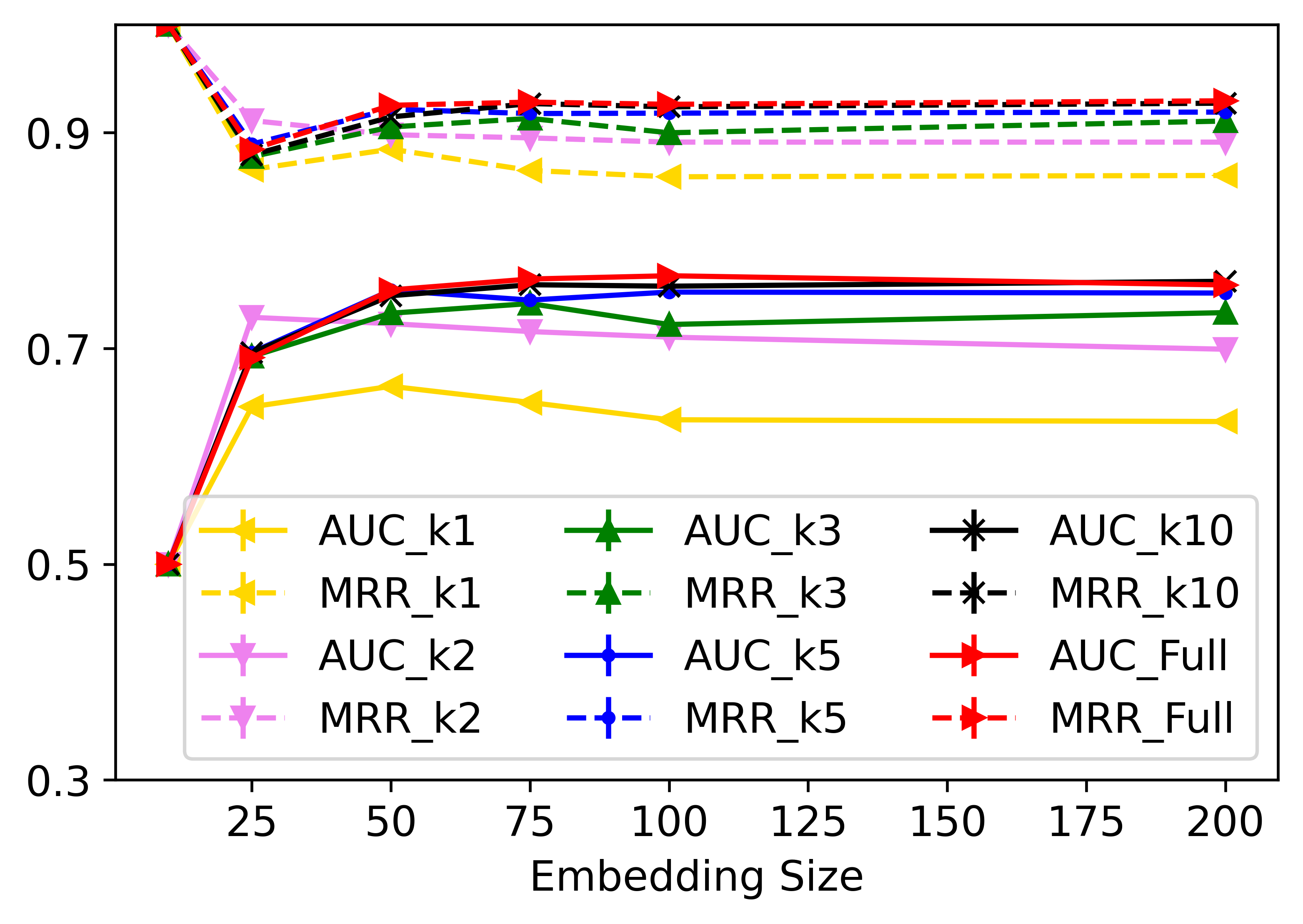}
        \vspace{-0.2in}
        \caption{\small \emph{ComplEx}}
        \label{fig:ComplEx}
     \end{subfigure}
     \hspace{\fill}
     \begin{subfigure}[t]{0.32\linewidth}
        \centering\includegraphics[width=\linewidth]{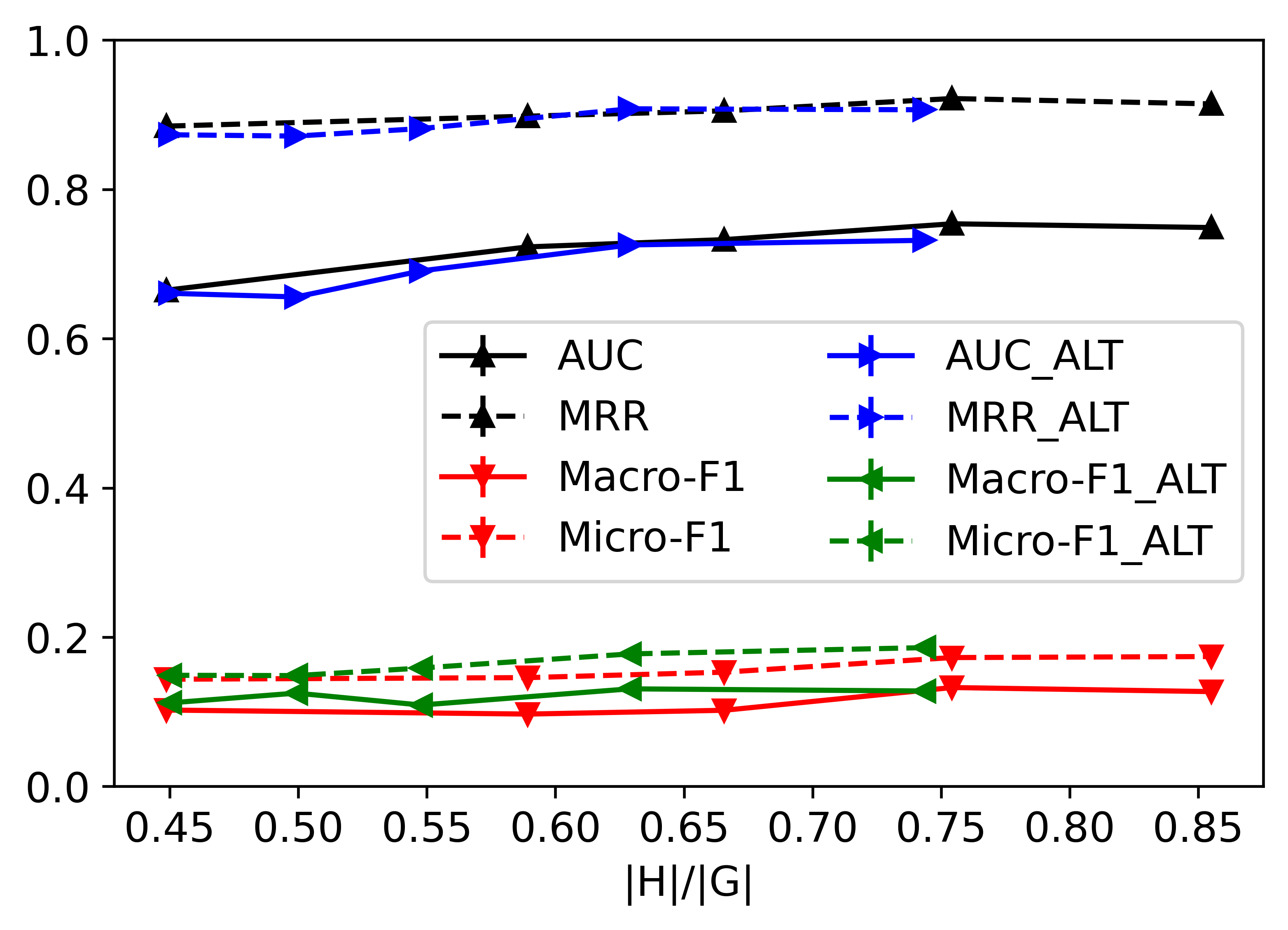}
        \vspace{-0.2in}
        \caption{\small Performance of \emph{ComplEx} in different sparsifiers in \emph{PubMed}}
        \label{fig:alt}
     \end{subfigure}
     \hspace{\fill}
     \begin{subfigure}[t]{0.32\linewidth}
        \centering\includegraphics[width=\linewidth]{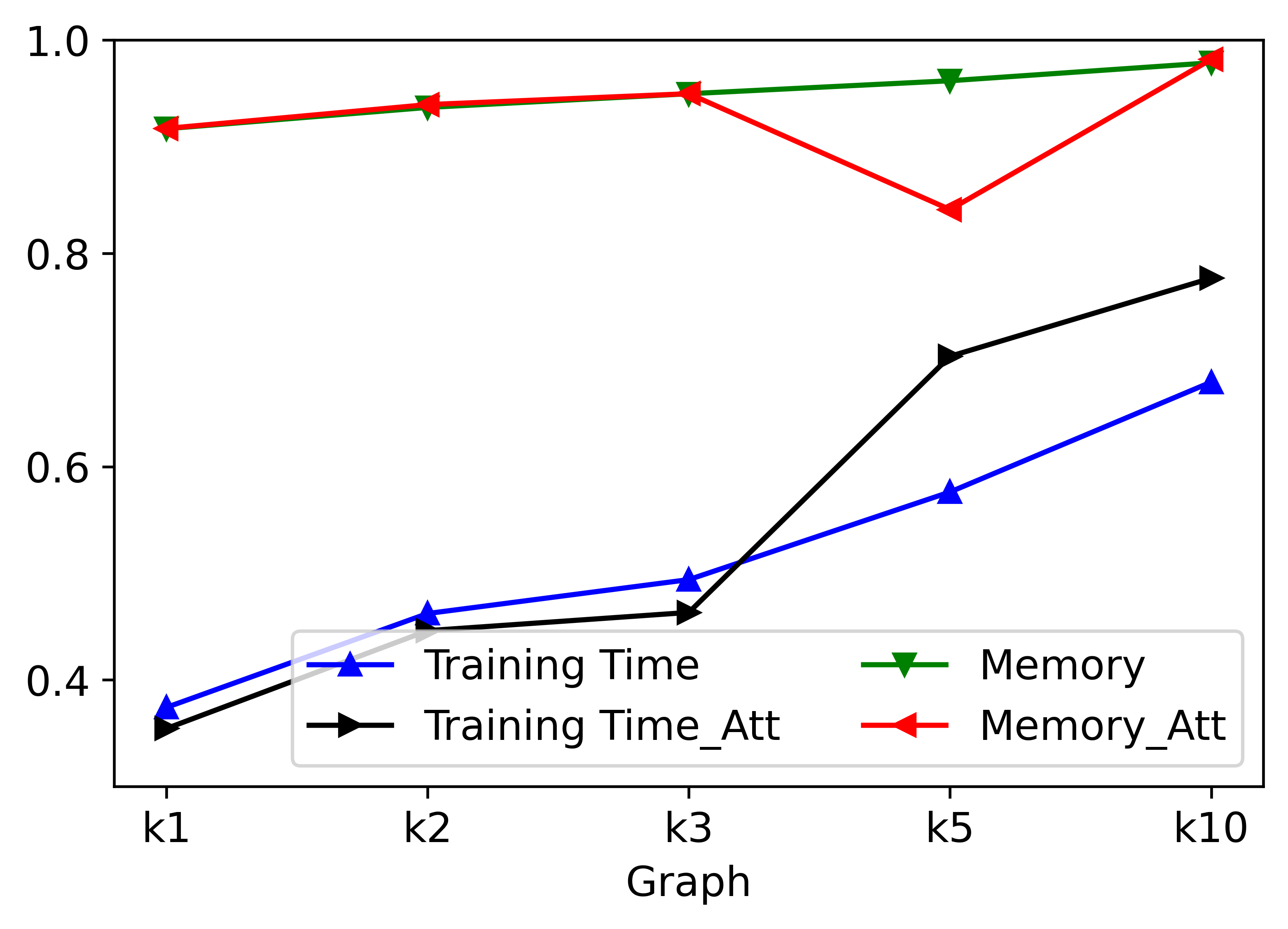}
        \vspace{-0.2in}
        \caption{\small Time and Memory in Attributed \emph{HAN}}
        \label{fig:att-han-tr}
     \end{subfigure}
     \hspace{\fill}
     \begin{subfigure}[t]{0.32\linewidth}
        \centering\includegraphics[width=\linewidth]{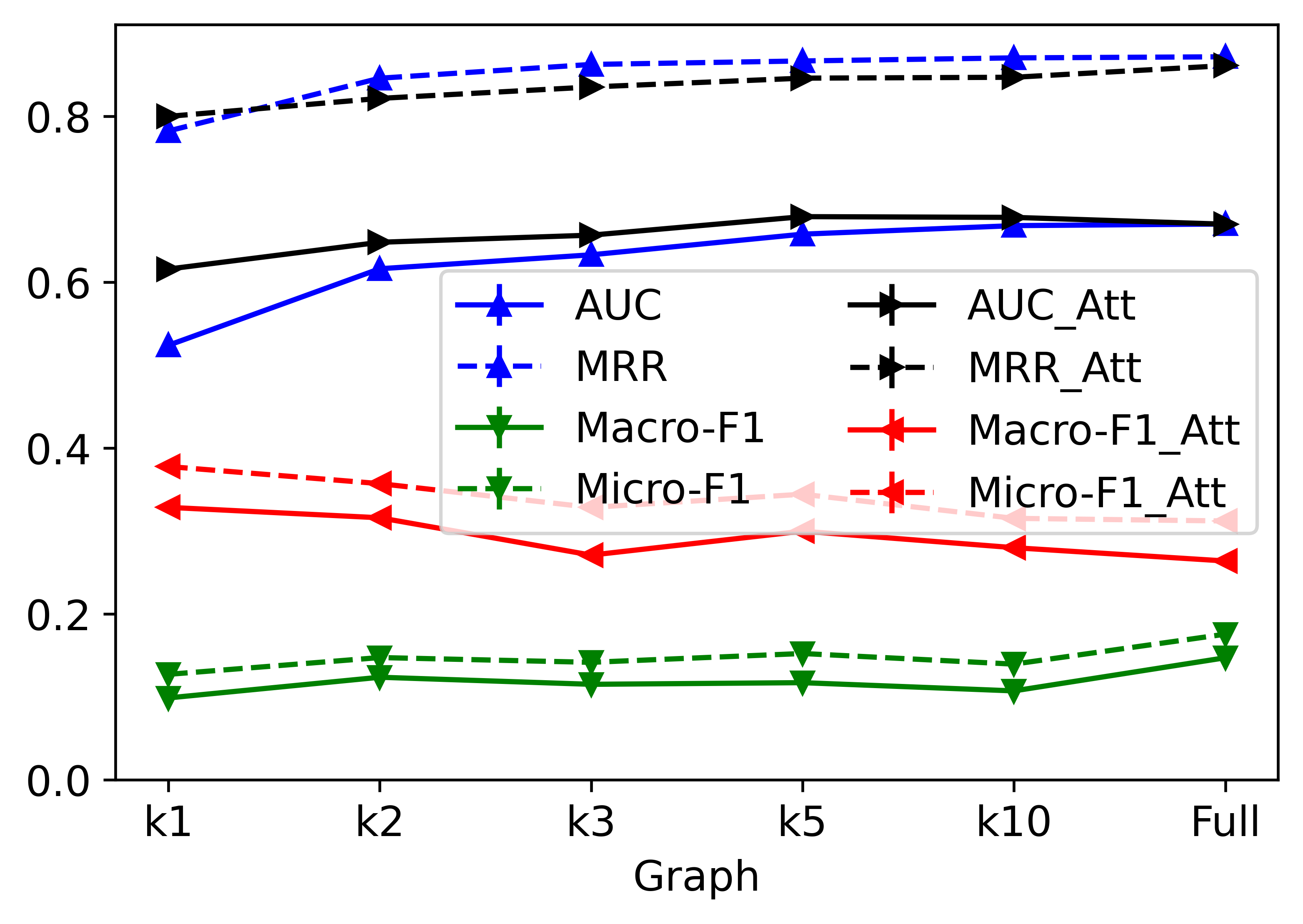}
        \vspace{-0.2in}
        \caption{\small \emph{AUC} \& \emph{MRR} in Attributed \emph{HAN}}
        \label{fig:att-han}
     \end{subfigure}
     \hspace{\fill}
     \begin{subfigure}[t]{0.32\linewidth}
        \centering\includegraphics[width=\linewidth]{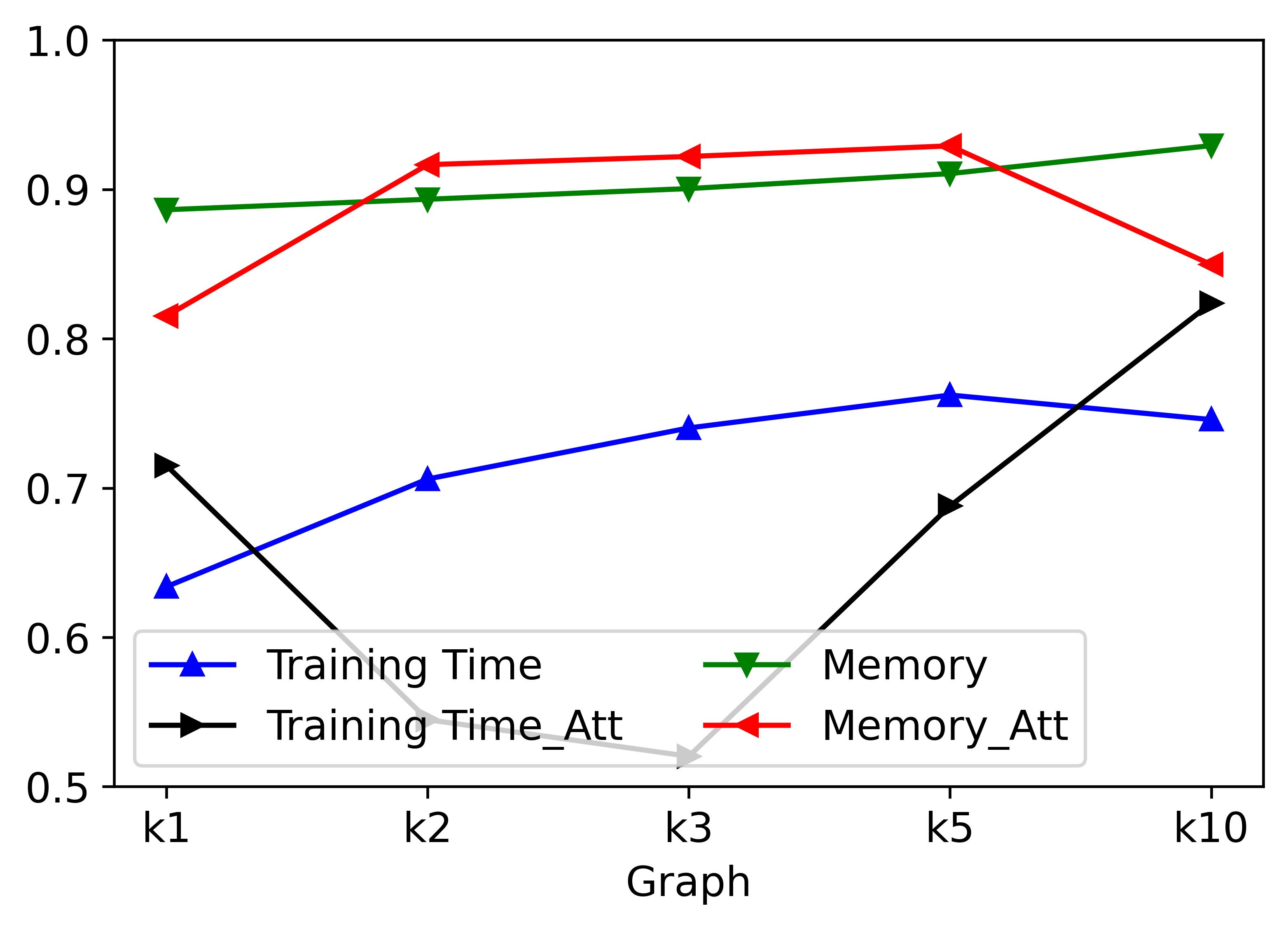}
        \vspace{-0.2in}
        \caption{\small Time and Memory in Attributed \emph{HGT}}
        \label{fig:att-hgt-tr}
     \end{subfigure}
     \hspace{0.2in}
     \begin{subfigure}[t]{0.32\linewidth}
        \centering\includegraphics[width=\linewidth]{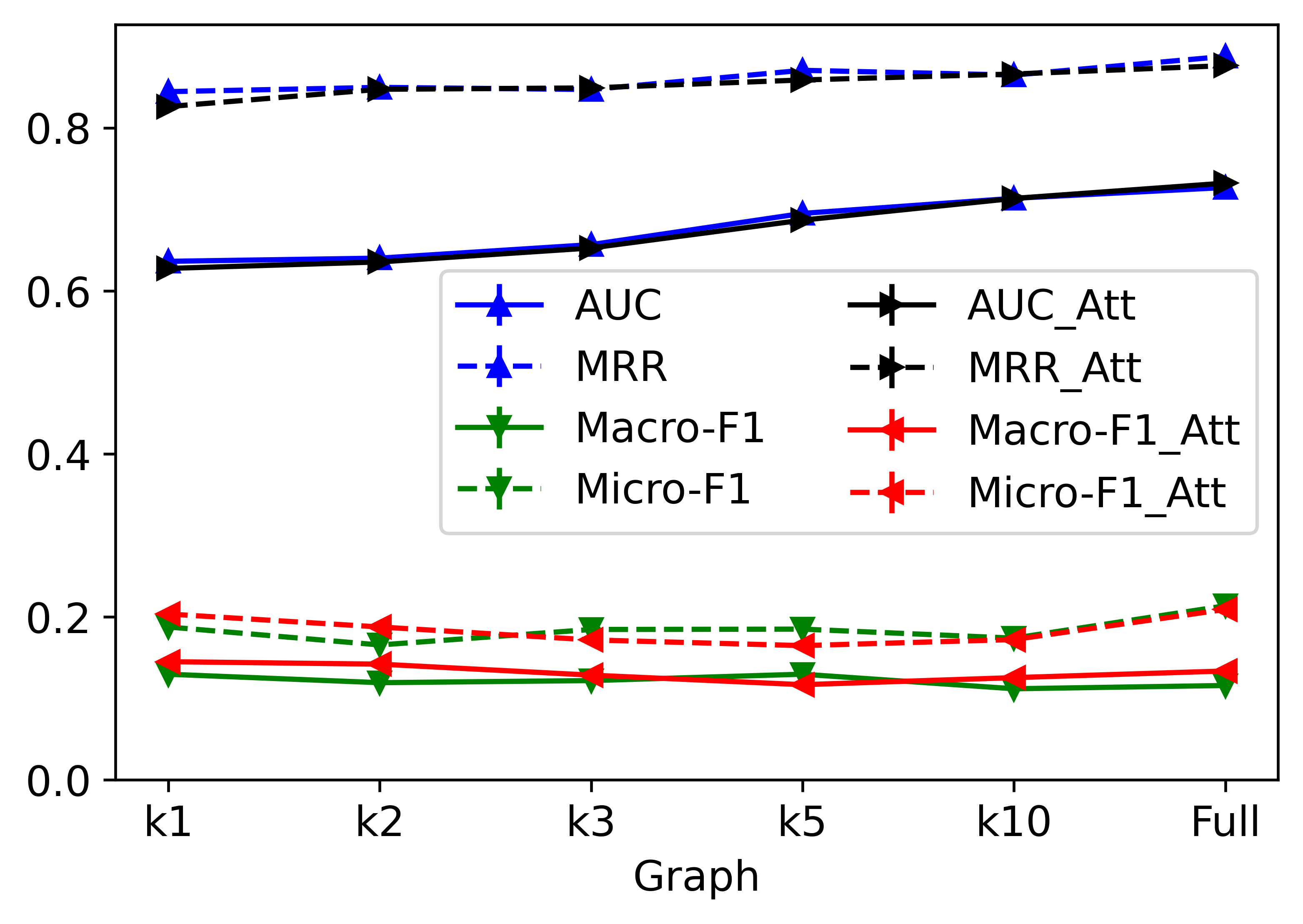}
        \vspace{-0.2in}
        \caption{\small \emph{AUC} \& \emph{MRR} in Attributed \emph{HGT}}
        \label{fig:att-hgt}
     \end{subfigure}
     \caption{\small Results of Extra Sets of Experiments Under the \emph{PubMed} Dataset}
     \label{fig:extra}
     \vskip -0.1in
\end{figure*}

{\noindent \bf Baseline Setting.}
In the baseline setting, we focus on unattributed representation learning that does not exploit node attributes and use the embedding size $50$. We first show effectiveness of the developed algorithm in reducing the size of heterogeneous graphs, and then compare performance of learning tasks in the sparsified graphs and the full graph. Table \ref{tab:sparsified} shows the \emph{sparsification ratio}, \emph{i.e.}, the ratio of the size of a sparsifier $H$ and the size of the full graph $G$, for varied values of $k$. It can be seen that our method consistently reduces the number of edges in the sparsifiers. The reduction becomes small with an increase in the value of $k$ since more edges are kept in the sparsifiers then. In general, the empirical performance matches our theory that the constructed sparsifiers have size $O(ktn)$.

Figure \ref{fig:results} plots performance of link prediction and node classification when applying different representation learning algorithms in the sparsified graphs and the full graph\footnote{HAN reported out-of-memory errors in \emph{Yelp} and \emph{Freebase} and \emph{AspEm} did not complete within 2 weeks in \emph{Freebase}.}. The training times for the sparsifiers in the \emph{PubMed} dataset are significantly improved to lower than $70\%$ of the training time required in the full graph except for \emph{AspEm}, as shown in Figure \ref{fig:time-pubmed}. The training times for \emph{HIN2Vec} even drop to lower than $44\%$ for all values of $k$. Interestingly, the reductions in training times are \emph{larger} than the sparsification ratio in Table \ref{tab:sparsified}, \emph{e.g.}, $85\%$ when $k=10$. It means that the reductions in computational time are remarkably larger than those in the graph size. The reductions in larger graphs such as \emph{Yelp} are even greater and drop to lower than $60\%$ of the original time. The running time of \emph{ComplEx} are consistently lower than $40\%$ while maintaining a good performance as will be shown later. The improvement in memory consumption during model training is not that obvious as the training time in Figure \ref{fig:ram-pubmed}. This may be due to that the additional overhead in the training excluding loading the graph data is comparable to the size of \emph{PubMed}. But the improvement of memory consumption in larger graphs \emph{Yelp} and \emph{Freebase} become much clearer as in Figures \ref{fig:ram-yelp} and \ref{fig:ram-freebase} and can be lower than $70\%$ of the original consumption.

Importantly, as in Figures \ref{fig:auc-pubmed}, \ref{fig:auc-yelp}, and \ref{fig:auc-freebase}, the performance \emph{AUC} and \emph{MRR} of link prediction are not significantly affected by the sparsification, especially for large values of $k$. Interestingly, the performance of \emph{ComplEx} and \emph{HIN2Vec} for quite a few settings can be even \emph{better} than that based on the full graph. We analyze that it could be because noises in the original graph are also removed during the sparsifying process. Among the tested embedding algorithms, \emph{ComplEx} consistently achieved the highest \emph{AUC} and \emph{MRR} values in different datasets. In Figures \ref{fig:f1-pubmed}, \ref{fig:f1-yelp}, and \ref{fig:f1-freebase}, the \emph{F1} values are generally comparable to the values in the full graph, and even better in some settings. Furthermore, graph (edge) sparsification appears to have more impacts on the performance of link prediction than on node classification. This seems to suggest that the task of node classification depends less on the edge connectivity compared to link prediction.


Due to superior performance of \emph{ComplEx}, we choose it as the default embedding model for comparing Alg. \ref{alg:graph} and \emph{ALT}. Figure \ref{fig:alt} plots the performance of \emph{ComplEx} based on the two types of sparsified graphs with respect to different sparsification ratios. It can be seen that \emph{AUC} and \emph{MRR} values of Alg. \ref{alg:graph} are slightly better than those of \emph{ALT}, while both \emph{F1} values of \emph{ALT} are larger by a small margin. In practice, users may select a sparsification method that is most suitable for their target application.

{\noindent \bf Additional Experiments.}
We first vary the size of the learned embedding from $10$ to $200$ ($[10, 25, 50, 75, 100, 200]$) in \emph{PubMed}. We observe that improvements in the training time and memory consumption are always significant, similar to the baseline setting. As expected, with an increase in the embedding size, the \emph{AUC} and \emph{MRR} are gradually improved due to a better learned embedding, as in Figures \ref{fig:HAN}, \ref{fig:HIN2Vec} and \ref{fig:ComplEx}. For \emph{HIN2Vec}, multiple sparsifies achieve higher \emph{AUC}/\emph{MRR} than the full graph, those red lines in Figure \ref{fig:HIN2Vec}. For \emph{HAN} and \emph{ComplEx}, the performance of the sparsifies with larger values of $k$, \emph{e.g.}, $k=10$, are comparable to that of the full graph as shown in Figures \ref{fig:HAN} and \ref{fig:ComplEx}.

Furthermore, since \emph{HAN} and \emph{HGT} can incorporate node attributes into the embedding learning, we perform attributed model training by taking node attributes into consideration. The training times and memory consumption in the attributed setting become slightly larger due to the processing of more information. But they are still much lower than training times and memory consumption in the full graph, as shown in Figures \ref{fig:att-han-tr} and \ref{fig:att-hgt-tr}. For \emph{HAN}, the accuracy performance in attributed graphs (red and black lines) are better than in unattributed graphs, especially for \emph{F1} values as shown in Figure \ref{fig:att-han}. \emph{HGT} appears to be strong in exploiting graph connectvitiy to generate accurate prediction even without the attribute information, as in Figure \ref{fig:att-hgt}. Consistent with the baseline setting, sparsification always brings great benefits in reducing computational resources while producing good accuracy performance.

\section{Conclusion and Future Work}
In this work, we study the problem of sparsifying heterogeneous graphs while well approximating the original graphs for subsequent graph learning tasks. We develop an efficient sampling method which takes the distribution of different types of edges into consideration and perform computational and storage complexity analysis. Extensive experiments validate the efficiencies of our proposed algorithms while attaining promising performance in downstream graph learning tasks. As the future work, we will study sophisticated algorithms for sparsifying heterogeneous graphs, \emph{e.g.}, by sampling edges according to edge (betweenness) centrality measures and keeping only important edges with high centrality. It is also an interesting direction to develop deep learning models to perform end-to-end heterogeneous graph sparsification and graph learning tasks.

\section*{Acknowledgment}

We thank graduate students Chandra Malgari and Divya Sanga for their help in the experiments and anonymous reviewers for their constructive comments.

\bibliographystyle{abbrv}
\bibliography{IEEE-het-spa}

\begin{thebibliography}{10}

\bibitem{BK96}
A.~Benczur and D.~Karger.
\newblock {Approximating $s$-$t$ minimum cuts in $\tilde{O}(n^2)$ time}.
\newblock In {\em Proceedings of STOC Conference}, pages 47--55, 1996.

\bibitem{bordes2013translating}
A.~Bordes, N.~Usunier, A.~Garcia-Duran, J.~Weston, and O.~Yakhnenko.
\newblock Translating embeddings for modeling multi-relational data.
\newblock {\em Advances in neural information processing systems}, 26, 2013.

\bibitem{CKL+18}
D.~Calandriello, I.~Koutis, A.~Lazaric, and M.~Valko.
\newblock {Improved large-scale graph learning through ridge spectral
  sparsification}.
\newblock In {\em Proceedings of ICML Conference}, pages 688--697, 2018.

\bibitem{dong2017metapath2vec}
Y.~Dong, N.~V. Chawla, and A.~Swami.
\newblock metapath2vec: Scalable representation learning for heterogeneous
  networks.
\newblock In {\em Proceedings of the 23rd ACM SIGKDD international conference
  on knowledge discovery and data mining}, pages 135--144, 2017.

\bibitem{fan2008liblinear}
R.-E. Fan, K.-W. Chang, C.-J. Hsieh, X.-R. Wang, and C.-J. Lin.
\newblock Liblinear: A library for large linear classification.
\newblock {\em the Journal of machine Learning research}, 9:1871--1874, 2008.

\bibitem{fan2019metapath}
S.~Fan, J.~Zhu, X.~Han, C.~Shi, L.~Hu, B.~Ma, and Y.~Li.
\newblock Metapath-guided heterogeneous graph neural network for intent
  recommendation.
\newblock In {\em Proceedings of the 25th ACM SIGKDD International Conference
  on Knowledge Discovery \& Data Mining}, pages 2478--2486, 2019.

\bibitem{fu2017hin2vec}
T.-Y. Fu, W.-C. Lee, and Z.~Lei.
\newblock Hin2vec: Explore meta-paths in heterogeneous information networks for
  representation learning.
\newblock In {\em Proceedings of the 2017 ACM on Conference on Information and
  Knowledge Management}, pages 1797--1806, 2017.

\bibitem{FHH+11}
W.~Fung, R.~Hariharan, N.~J. Harvey, and D.~Panigrahi.
\newblock {A general framework for graph sparsification}.
\newblock In {\em Proceedings of STOC Conference}, pages 71--80, 2011.

\bibitem{hu2020heterogeneous}
Z.~Hu, Y.~Dong, K.~Wang, and Y.~Sun.
\newblock Heterogeneous graph transformer.
\newblock In {\em Proceedings of The Web Conference 2020}, pages 2704--2710,
  2020.

\bibitem{jiang2021pre}
X.~Jiang, T.~Jia, Y.~Fang, C.~Shi, Z.~Lin, and H.~Wang.
\newblock Pre-training on large-scale heterogeneous graph.
\newblock In {\em Proceedings of the 27th ACM SIGKDD Conference on Knowledge
  Discovery \& Data Mining}, pages 756--766, 2021.

\bibitem{nicholson2020constructing}
D.~N. Nicholson and C.~S. Greene.
\newblock Constructing knowledge graphs and their biomedical applications.
\newblock {\em Computational and structural biotechnology journal},
  18:1414--1428, 2020.

\bibitem{PS89}
D.~Peleg and A.~Schaffer.
\newblock {Graph spanners}.
\newblock {\em Journal of Graph Theory}, 13(1):99--116, 1989.

\bibitem{SWT16}
V.~Sadhanala, Y.-X. Wang, and R.~J. Tibshirani.
\newblock {Graph sparsification approaches for Laplacian smoothing}.
\newblock In {\em Proceedings of AISTATS Conference}, pages 1250--1259, 2016.

\bibitem{shi2018aspem}
Y.~Shi, H.~Gui, Q.~Zhu, L.~Kaplan, and J.~Han.
\newblock Aspem: Embedding learning by aspects in heterogeneous information
  networks.
\newblock In {\em Proceedings of the 2018 SIAM International Conference on Data
  Mining}, pages 144--152. SIAM, 2018.

\bibitem{SS11}
D.~Spielman and N.~Srivastava.
\newblock {Graph sparsification by effective resistances}.
\newblock {\em SIAM Journal on Computing}, 40(6):1913--1926, 2011.

\bibitem{ST11}
D.~Spielman and S.-H. Teng.
\newblock {Spectral sparsification of graphs}.
\newblock {\em SIAM Journal on Computing}, 40(4):981--1025, 2011.

\bibitem{sun2019rotate}
Z.~Sun, Z.-H. Deng, J.-Y. Nie, and J.~Tang.
\newblock Rotate: Knowledge graph embedding by relational rotation in complex
  space.
\newblock {\em arXiv preprint arXiv:1902.10197}, 2019.

\bibitem{trouillon2016complex}
T.~Trouillon, J.~Welbl, S.~Riedel, {\'E}.~Gaussier, and G.~Bouchard.
\newblock Complex embeddings for simple link prediction.
\newblock In {\em International conference on machine learning}, pages
  2071--2080. PMLR, 2016.

\bibitem{wang2019heterogeneous}
X.~Wang, H.~Ji, C.~Shi, B.~Wang, Y.~Ye, P.~Cui, and P.~S. Yu.
\newblock Heterogeneous graph attention network.
\newblock In {\em The World Wide Web Conference}, pages 2022--2032, 2019.

\bibitem{yang2020heterogeneous}
C.~Yang, Y.~Xiao, Y.~Zhang, Y.~Sun, and J.~Han.
\newblock Heterogeneous network representation learning: A unified framework
  with survey and benchmark.
\newblock {\em IEEE Transactions on Knowledge and Data Engineering}, 2020.

\bibitem{zhang2019heterogeneous}
C.~Zhang, D.~Song, C.~Huang, A.~Swami, and N.~V. Chawla.
\newblock Heterogeneous graph neural network.
\newblock In {\em Proceedings of the 25th ACM SIGKDD International Conference
  on Knowledge Discovery \& Data Mining}, pages 793--803, 2019.

\bibitem{ZLB21b}
C.~Zhu, Q.~Liu, and J.~Bi.
\newblock {Communication Efficient Distributed Hypergraph Clustering}.
\newblock In {\em Proceedings of the 44th International ACM SIGIR Conference on
  Research and Development in Information Retrieval (SIGIR)}, pages 2131--2135,
  2021.

\bibitem{ZLB21a}
C.~Zhu, Q.~Liu, and J.~Bi.
\newblock {Spectral Vertex Sparsifiers and Pair-Wise Spanners over Distributed
  Graphs}.
\newblock In {\em Proceedings of ICML Conference}, pages 12890--12900, 2021.

\bibitem{ZSL+19a}
C.~Zhu, S.~Storandt, K.-Y. Lam, S.~Han, and J.~Bi.
\newblock {Improved dynamic graph learning through fault-tolerant
  sparsification}.
\newblock In {\em Proceedings of ICML Conference}, pages 7624--7633, 2019.

\end{thebibliography}

\end{document}